\documentclass[review, 3p, sort&compress]{elsarticle}

\usepackage{lineno,hyperref}
\modulolinenumbers[5]

\graphicspath{{Images/}}
\DeclareGraphicsExtensions{.pdf,.jpeg,.png,.jpg,.svg}

\usepackage{amsmath}
\usepackage{amssymb}
\usepackage{textcomp}
\usepackage{makecell}
\usepackage{multirow}
\usepackage{multicol}
\usepackage{url}
\usepackage{xcolor}

\journal{Pattern Recognition}

\begin{document}

\begin{frontmatter}

\title{Semantic-Aware Scene Recognition}

\author[1]{Alejandro L\'{o}pez-Cifuentes\corref{cor1}}
\ead{alejandro.lopezc@uam.es}
\cortext[cor1]{Corresponding author}
\author[1]{Marcos Escudero-Vi\~nolo}
\ead{marcos.escudero@uam.es}
\author[1]{Jes\'{u}s~Besc\'{o}s}
\ead{j.bescos@uam.es}
\author[1]{\'{A}lvaro Garc\'{i}a-Mart\'{i}n}
\ead{alvaro.garcia@uam.es}

\address[1]{Video Processing and Understanding Lab, Universidad Aut\'{o}noma de Madrid, 28049, Madrid, Spain}

\begin{abstract}

Scene recognition is currently one of the top-challenging research fields in computer vision. This may be due to the ambiguity between classes: images of several scene classes may share similar objects, which causes confusion among them. The problem is aggravated when images of a particular scene class are notably different. Convolutional Neural Networks (CNNs) have significantly boosted performance in scene recognition, albeit it is still far below from other recognition tasks (e.g., object or image recognition). In this paper, we describe a novel approach for scene recognition based on an end-to-end multi-modal CNN that combines image and context information by means of an attention module. Context information, in the shape of a semantic segmentation, is used to gate features extracted from the RGB image by leveraging on information encoded in the semantic representation: the set of scene objects and stuff, and their relative locations. This gating process reinforces the learning of indicative scene content and enhances scene disambiguation by refocusing the receptive fields of the CNN towards them. Experimental results on three publicly available datasets show that the proposed approach outperforms every other state-of-the-art method while significantly reducing the number of network parameters. All the code and data used along this paper is available at: \url{https://github.com/vpulab/Semantic-Aware-Scene-Recognition}
\end{abstract}


\end{frontmatter}


\section{Motivation and Problem Statement}\label{sec:Problem Motivation}

Scene recognition is a hot research topic whose complexity is, according to reported performances \cite{zhou2018places}, on top of image understanding challenges. The paradigm shift forced by the advent of Deep Learning methods, and specifically, of Convolutional Neural Networks (CNNs), has significantly enhanced results, albeit they are still far below those achieved in tasks as image classification, object detection and semantic segmentation \cite{ILSVRC15ImageNet,zhou2017scene,Cordts2016Cityscapes}.

\begin{figure}[t!]
    \centering
    \includegraphics[width=0.55\linewidth,keepaspectratio]{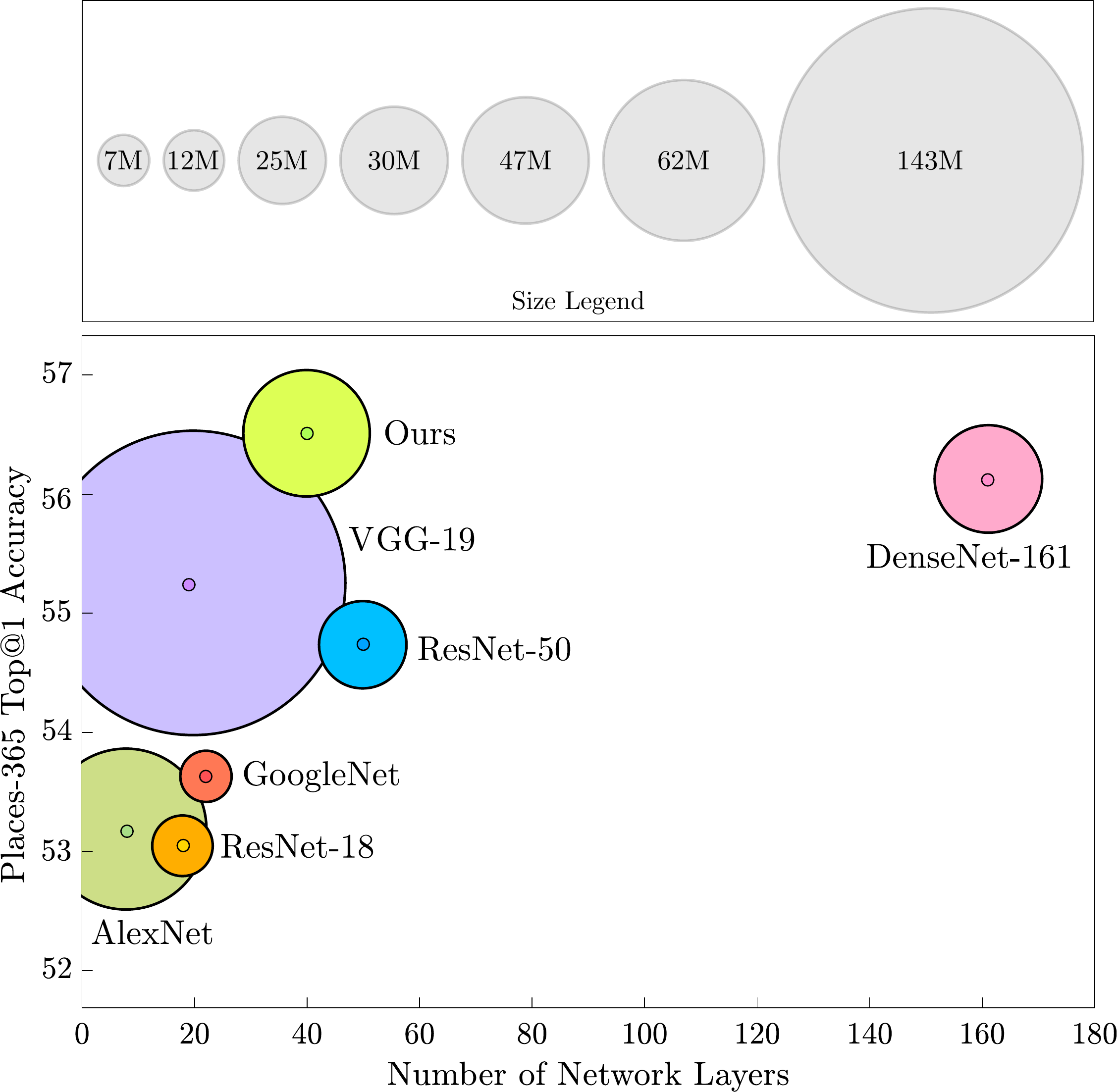}
    \caption{Scene recognition performance of CNN-based solutions on the Places-365 Standard Dataset \cite{zhou2018places}. Each network is represented by its Top@1 accuracy (vertical axis), its number of layers (horizontal axis) and its number of units (radius of each circle). Circles' colors are only for better identification.}
    \label{fig:Places Networks}
\end{figure}

The complexity of the scene recognition task lies partially on the ambiguity between different scene categories showing similar appearances and objects' distributions: inter-scene boundaries can be blurry, as the sets of objects that define a scene might be highly similar to another's. Therefore, scene recognition should cope with the classical tug-off-war between repeatable (handle intra-class variation) and distinctive (discriminate among different categories) characterizations. Whereas CNNs have been proved to automatically yield trade-off solutions with success, the complexity of the problem increases with the number of categories, and specially when training datasets are unbalanced.

Recent studies on the semantic interpretability of CNNs, suggest that the learning of scenes is inherent to the learning of the objects they include \cite{bau2017network}. Object \textit{detectors} somehow arise as latent-variables in hidden-units within networks trained to recognize scenes. These \textit{detectors} operate without constraining the networks to decompose the scene recognition problem \cite{zhou2014object}. The number of \textit{detectors} is, to some extent, larger for deeper and wider (with a larger number of units) networks. During the last years, the main trend to enhance scene recognition performance has focused on increasing the number of CNN units. However, performance does not increase linearly with the increase in the number of network parameters \cite{zhou2018places}. For instance, DenseNet-161 is twenty-times deeper than AlexNet, and VGG-16 has twenty-times more units than GoogleNet, but performances of DenseNet-161 and VGG-16 for scene recognition are only a \(2.9 \%\) and a \(1.6 \%\) better than those of AlexNet and GoogleNet, respectively (see Figure \ref{fig:Places Networks}).

\begin{figure*}[t!]
    \centering
    \includegraphics[width=1\linewidth,keepaspectratio]{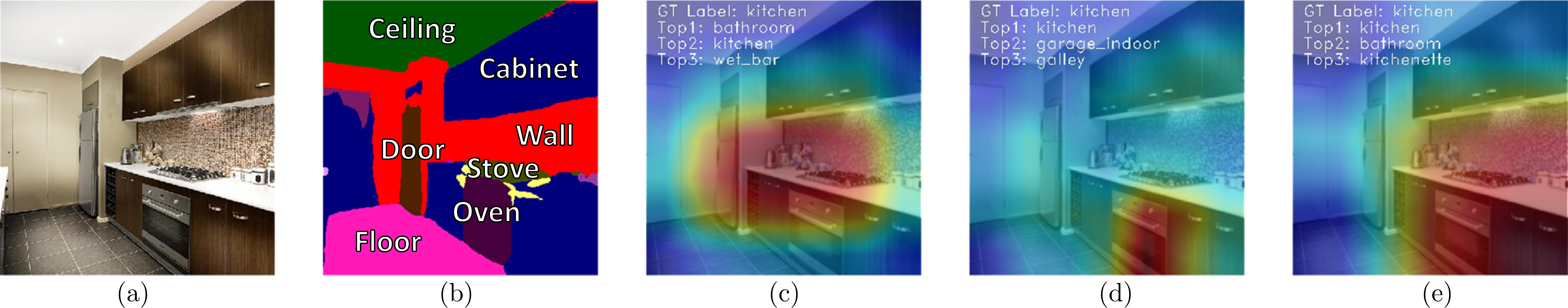}
    \caption{Scene recognition results depending on the input data. (a) RGB image corresponding to the "Kitchen" scene class. (b) Semantic segmentation from (a) obtained by a state-of-the-art CNN-based algorithm. (c) Class Activation Map (CAM) \cite{zhou2015cnnlocalization} just using the (a) RGB image. (d) CAM just using the (b) semantic segmentation. (e) CAM for the proposed approach, using both (a) and (b). Top@3 predicted classes are included in the top-left corner of images (c) to (e). Better viewed in color.}
    \label{fig:Focusing Example}
\end{figure*}

This paper presents a novel strategy to improve scene recognition: to use object-level information to guide scene learning during the training process. It is well-known that\textemdash probably due to the ImageNet fine-tuning paradigm \cite{he2018rethinking}, widely used models are biased towards the spatial center of the visual representation (see Figure \ref{fig:Focusing Example}) \cite{das2017human}. The proposed approach relies on semantic-driven attention mechanisms to prioritize the learning of common scene objects. This strategy achieves the best reported results on the validation set of the Places365 Standard dataset \cite{zhou2018places} (see Figure \ref{fig:Places Networks}) without increasing the network's width or deepness, i.e. using \(67.13 \%\) less units and \(62.73 \%\) less layers than VGG-16 and DenseNet-161 respectively.

Similar strategies have been recently proposed to constrain scene recognition through histograms of patch-wise object detections \cite{wang2017weakly,cheng2018scene, jiang2019deep}. Compared to these strategies, the proposed method naturally exploits the spatial relationships between objects and emphasizes interpretability by relying on semantic segmentation. Semantic segmentation can be understood as a dense object detection task. In fact, scene recognition, object detection and semantic segmentation are interrelated tasks that share a common Branch in the recently proposed \textit{taskonomies} or task-similarity branches \cite{zamir2018taskonomy}.  However, the performance achieved by semantic segmentation methods is generally lower than that of object detection ones mainly due to the difference in size and variety of their respective datasets. Howbeit, the proposed method surmounts this gap and yields a higher scene recognition performance than object-constrained methods \cite{wang2017weakly,cheng2018scene}, using a substantially smaller number of units (see Section \ref{sec:Experiments and Results}).

In essence, the proposed work aims to enhance scene recognition without increasing network's capacity or depth leading to lower training and inference times, as well as to smaller data requirements and training resources. The specific contributions of these paper are threefold:
\begin{enumerate}
  \item We propose an end-to-end multi-modal deep learning architecture which gathers both image and context information using a two-branched CNN architecture.
  \item We propose to use semantic segmentation as an additional information source to automatically create, through a convolutional neural network, an attention model to reinforce the learning of relevant contextual information.
  \item We validate the effectiveness of the proposed method yielding state-of-the-art results for the MIT Indoor 67 \cite{quattoni2009recognizing}, SUN 397 \cite{xiao2010sun} and Places365 \cite{zhou2018places} datasets.
\end{enumerate}

\section{Related Works} \label{sec:State of the Art}

\subsection{Scene Recognition}
A variety of works for scene recognition have been proposed during the last years: it is a hot topic. Existing methods can be broadly organized into two different categories: those based on hand-crafted feature representations and those based on CNN architectures. 

Among the first group, earliest works proposed to design a holistic feature representations. Generalized Search Trees (GIST) \cite{oliva2005gist} were used to generate a holistic low dimensional representation for each image. However, precisely due to the holistic approach, GIST lacks of scene's local structure information. To overcome this problem, local feature representations were used to exploit the pattern of local patches and combine their representations in a concatenated feature vector. Census Transform Histogram (CENTRIST) \cite{wu2011centrist} encodes the local structural properties within an image and suppresses detailed textural information to boost scene recognition performance. To further increase generalization, Oriented Texture Curves (OTC) \cite{margolin2014otc} captures patch textures along multiple orientations to be robust to illumination changes, geometric distortions and local contrast differences. Overall, despite reporting noteworthy results, these hand-crafted features, either holistic or local, exploit low-level features which may not be discriminative enough for ambiguous or highly related scenes \cite{zhou2018places}. Besides, the hand-craft nature of features may hinder their scalability as \textit{ad hoc} designs may be required for new domains.

Solutions based on CNNs generally result in higher performances. Recent network architectures, together with multi-million datasets such as Places-365, crush the results obtained by hand-crafted features \cite{zhou2018places}. CNNs exploit multi-scale feature representations using convolutional layers and do not require the design of hand-crafted features as they are implicitly learned through the training process. Besides, CNNs combine low-level latent information such as color, texture and material with high-level information, e.g., parts and objects to obtain better scene representations and boost scene recognition performance \cite{bau2017network}. Well-consolidated network architectures, such as AlexNet, GoogLeNet, VGG-16, ResNet-152 and DenseNet-161 have reported accuracies of \(53.17\%\), \(53.63\%\), \(55.24\%\), \(54.74\%\) and \(56.10\%\) respectively when classifying images from the challenging Places-365 Standard dataset \cite{zhou2018places}.

Features extracted from these networks have been also combined with different scene classification methods. For instance, Xie \textit{et al.} \cite{xie2015hybrid} proposed to use features from AlexNet and VGG with Mid-level Local Representation (MLR) and Convolutional Fisher Vector representation (CFV) dictionaries to incorporate local and structural information. Cimpoi \textit{et al.} \cite{cimpoi2015deep} used material and texture information extracted from a CNN-filter bank and from a bag-of-visual-words to increase generalization for enhanced scene recognition. Yoo \textit{et al.} \cite{yoo2014fisher} benefited from multi-scale CNN-based activations aggregated by a Fisher kernel framework to perform significantly better on the MIT Indoor Dataset \cite{quattoni2009recognizing}.

While accuracies from this networks and methods are much higher than those reported by methods based on hand-crafted features, they are far from those achieved by CNNs in other image classification problems (e.g., ImageNet challenge best reported accuracy is \(85.4\%\) \cite{mahajan2018exploring} while best reported accuracy on Person Re-Identification task is \(95.4\%\) \cite{quan2019auto}). 

As described in Section \ref{sec:Problem Motivation}, the increment in capacity (number of units) of CNNs does not, for scene recognition, lead to a linear rise in performance. This might be explained by the inability of the networks to handle inter-class similarities \cite{cheng2018scene}: unrelated scene classes may share objects that are prone to produce alike scene representations weakening the network's generalization power. To cope with this problem, some recent methods incorporate context and discriminative object information to constrain scene recognition. 

Based on context information, Xie \textit{et al.} \cite{xie2017lg} proposed to enhance fine-grained recognition by detecting part candidates based on saliency detection and by constructing a CNN architecture with local parts and global discrimination. Zhao \textit{et al.} \cite{zhao2018volcano}, similarly, proposed a discriminative discovery network (DisNet) that generates a discriminative map (Dis-Map) for the input image. This map is further used to select scale-aware discriminative locations which are then forwarded to a multi-scale pipeline for CNN feature extraction.

In terms of discriminative object information, Herranz-Perdiguero \textit{et al.} \cite{herranz2018pixels} proposed an extension of the DeepLab semantic segmentation network by introducing SVMs classifiers to perform scene recognition based on object histograms. In the same vein, Wang \textit{et al.} \cite{wang2017weakly} designed an architecture where patch-based features are extracted from customized object-wise and scene-wise CNNs to construct semantic representations of the scene. A scene recognition method built on these representations (Vectors of Semantically Aggregated Descriptors (VSAD)) yields excellent performance on standard scene recognition benchmarks. VSAD's performance has been recently enhanced by measuring correlations between objects among different scene classes \cite{cheng2018scene}. These correlations are then used to reduce the effect of common objects in scene miss-classification and enhance the effect of discriminative objects through a Semantic Descriptor with Objectness (SDO).

Even though these methods constitute state-of-the-art in scene recognition, their dependency of object information, obtained by using patch-based object classification techniques, entails severe and reactive parametrization (scale, patch-size, stride, overlapping...). Moreover, their descriptors are object-centered and lack of information on the spatial-interrelations between object instances. We instead propose to use an end-to-end CNN that exploits spatial relationships by using semantic segmentation instead of object information to guide network's attention. This proves to provide equal or better performance while significantly reducing the number of network units and hyper-parameters.

\subsection{Semantic Segmentation}
Semantic segmentation is the task of assigning a unique object (or stuff) label to every pixel of an image. As reported in several benchmarks (MIT Scene Parsing Benchmark \cite{zhou2017scene}, CityScapes Dataset \cite{Cordts2016Cityscapes}, and COCO Challenge \cite{lin2014microsoftCOCO}), top-performing strategies, which are completely based on end-to-end deep learning architectures, are year by year getting closer to human accuracy.

Among the top performing strategies, Wang \textit{et al.} proposed to use a dense up-sampling CNN to generate pixel-level predictions within a hybrid dilated convolution framework \cite{wang2017understanding}. In this vein, the Unified Perceptual Parsing \cite{xiao2018unified} benefits from a multi-task framework to recognize several visual concepts given a single image (semantic segmentation, materials, texture). By regularization, when multiple tasks are trained simultaneously, their individual results are boosted. Zhao \textit{et al.} modelled contextual information between different semantic labels proposing the Pyramid Pooling Module \cite{zhao2016pyramid}\textemdash e.g., an airplane is likely to be on a runway or flying across the sky but rarely over the water. These spatial relationships allow reducing the complexity associated with large sets of object labels, generally improving performance. Guo \textit{et al.} \cite{guo2018small} proposed the Inter-Class Shared Boundary (ISB) Encoder which aims to encode the level of spatial adjacency between object-class pairs into the segmentation network to obtain higher accuracy for classes associated with small objects. 

Motivated by the success of modelling contextual information and spatial relationships, we decided to explore the use of semantic segmentation as an additional modality of information to confront scene recognition.

\subsection{Multi-Modal Deep Learning Architectures}
The use of complementary image modalities (e.g., depth, optical flow or heat maps) as additional sources of information to multi-modal classification models has been recently proposed to boost several computer vision tasks: pedestrian detection \cite{park2013exploring, yang2015convolutional, daniel2016semantic, mao2017can}, action recognition \cite{simonyan2014two, park2016combining} and hand-gesture recognition \cite{abavisani2018improving}.

Regarding pedestrian detection, a plethora of solutions have been proposed to include additional sources of information besides the RGB one \cite{park2013exploring, yang2015convolutional, daniel2016semantic, mao2017can}. Park \textit{et al.} proposed the incorporation of optical-flow-based features to model motion through a decision-forest detection scheme \cite{park2013exploring}. Convolutional Channel Features (CCF) \cite{yang2015convolutional} rely on extracted low-level features from VGG-16 network as an additional input channel to enhance pedestrian detection. Daniel \textit{et al.} \cite{daniel2016semantic} used short range and long range multiresolution channel features obtained by eight different individual classifiers to propose a detector, obtaining state-of-the-art results at a higher speed. An extensive study of the features that may help pedestrian detection has been carried out by Mao \textit{et al.} \cite{mao2017can}. Authors suggest that mono-modal pedestrian detection results are improved when depth, heat-maps, optical flow, or segmentation representations are integrated as additional information modalities into a CNN-based pedestrian detector.

The effectiveness of modelling temporal features using multi-modal networks has been also studied \cite{simonyan2014two, park2016combining}. Simonyan \textit{et al.} \cite{simonyan2014two} and Park \textit{et al.} \cite{park2016combining} proposed to incorporate optical flow as an additional input for action recognition. These methods lead to state-of-the-art performance when classifying actions in videos. Similarly, hand-gesture recognition has also benefited from the use of optical-flow \cite{abavisani2018improving}.

In this paper, we propose to combine RGB images and their corresponding semantic segmentation. Additionally, instead of using traditional combinations\textemdash linear combination, feature concatenation or averaging \cite{simonyan2014two, park2016combining,mao2017can}, we propose to use a CNN architecture based on a novel attention mechanism.

\subsection{Attention Networks}
CNNs with architectures based on attention modules have been recently shown to improve performances on a wide and varied set of tasks \cite{gregor2015draw}. These attention modules are inspired by saliency theories on the human visual system, and aim to enhance CNNs by refocusing them onto task-relevant image content, emphasizing the learning of this content \cite{woo2018cbam}.

For image classification, the Residual Attention Network \cite{wang2017residual}, based on encoder-decoder stacked attention modules, increases classification performance by generating attention-aware features. The Convolutional Block Attention Module (CBAM) \cite{woo2018cbam} infers both spatial and channel attention maps to refine features at all levels while obtaining top performance on ImageNet classification tasks and on MS COCO and VOC2007 object detection datasets. An evaluation benchmark proposed by Kiela \textit{et al.} \cite{kiela2018efficient}, suggested that the incorporation of different multi-modal attention methods, including additive, max-pooling, and bilinear models is highly beneficial for image classification algorithms. 

In terms of action recognition and video analysis, attention is used to focus on relevant content over time. Yang \textit{et al.} proposed to use soft attention models by combining a convolutional Long Short-Term Memory Network (LSTM) with a hierarchical architecture to improve action recognition metrics \cite{yan2017cham}. 

Guided by the broad range and impact of attention mechanisms, in this paper we propose to train an attention mechanism based on semantic information. For the semantic segmentation of a given image, this mechanism returns a map based on prior learning. This map promises to reinforce the learning of features derived from common scene objects and to hinder the effect of features produced by non-specific objects in the scene.

\section{Proposed Architecture}\label{sec:Proposed Method}
\begin{figure*}[t!]
    \centering
    \includegraphics[width=1\linewidth,keepaspectratio]{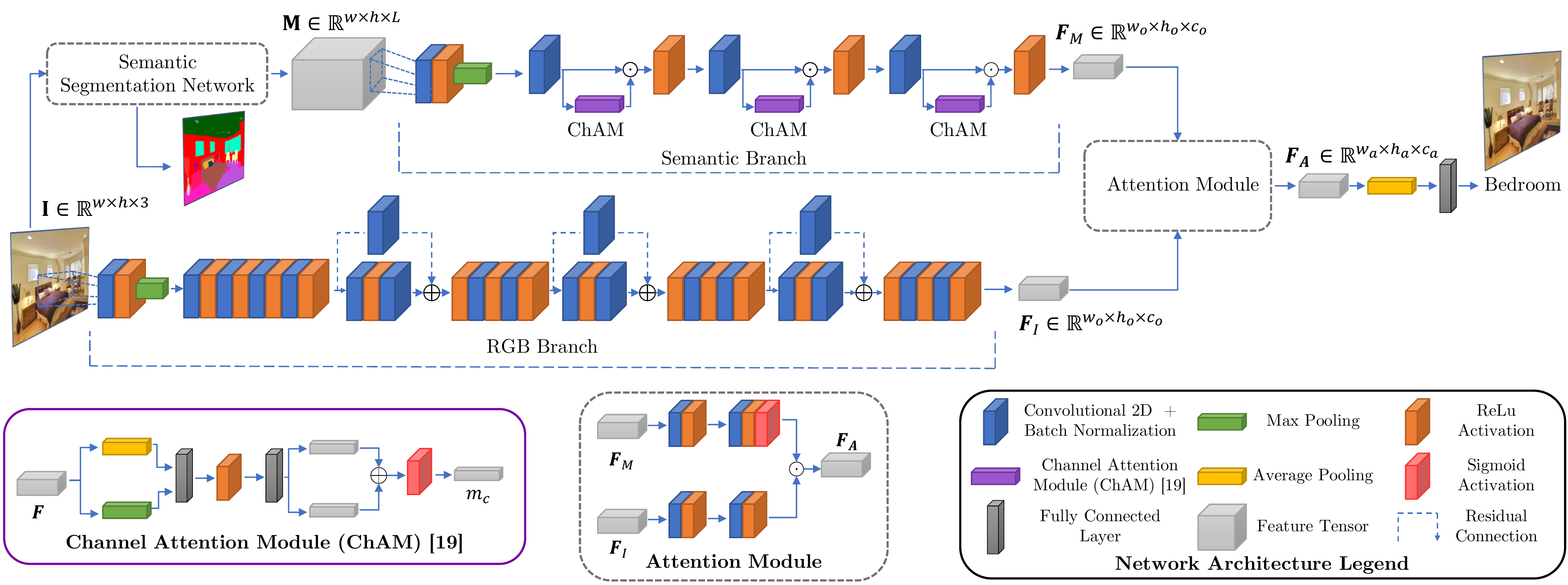}
    \caption{Architecture of the multi-modal deep-learning scene recognition model. The architecture is composed of a Semantic Branch, a RGB Branch and an Attention Module. The semantic branch aims to extract meaningful features from a semantic segmentation score map. This Branch aims to convey an attention map only based on meaningful and representative scene objects and their relationships. The RGB Branch extracts features from the color image. These features are then gated by the semantic-based attention map in the Attention Module. Through this process, the network is automatically refocused towards the meaningful objects learned as relevant for recognition by the Semantic Branch. Better viewed in color.}
    \label{fig:Network Architecture}
\end{figure*}

This paper proposes a multi-modal deep-learning scene recognition system composed of a two-branched CNN and an Attention Module (see Figure \ref{fig:Network Architecture}).

\subsection{Motivations}
The basic idea behind the design is that features extracted from RGB and semantic segmentation domains complement each other for scene recognition. If we train a CNN to recognize scenes just based on the semantic segmentation of an input RGB image (see top Branch in Figure \ref{fig:Network Architecture}), this network tends to focus on representative and discriminative scene objects, yielding genuine-semantic representations based on objects and stuff classes. We propose to use these representations to train an attention map. This map is then used to gate RGB-extracted representations (see bottom Branch in Figure \ref{fig:Network Architecture}), hence refocusing them on the learned scene content.

The effect of this attention mechanism is that the receptive field of the most probable neuron at the final classification layer\textemdash known as class-activation map \cite{zhou2015cnnlocalization}\textemdash is enlarged and displaced towards scene discriminative elements that have been learned by the Semantic Branch. An example of this behavior is depicted in Figure \ref{fig:Focusing Example}: the RGB Branch is focused on the center of the image; the Semantic Branch mainly focuses on the \textit{oven} for predicting the scene class; the combined network includes the \textit{oven} and other "Kitchen" discriminative objects as the \textit{stove} and the \textit{cabinet} for prediction.

\subsection{Preliminaries} \label{subsec:Network Architecture}
The proposed network architecture is presented in Figure \ref{fig:Network Architecture}. Let \(\textbf{I} \in \mathbb{R}^{w \times h \times 3}\) be a normalized RGB image and let \(\textbf{M} \in \mathbb{R}^{w \times h \times L}\) be a score tensor representing \(\textbf{I}\)'s semantic segmentation, where \(\textbf{M}_{i,j} \in \mathbb{R}^{1 \times 1 \times L}\) represents the probability distribution for the \(i,j\)-th pixel on the \(L\)-set of learned semantic labels. 

The network is composed of a Semantic Branch (Figure \ref{fig:Network Architecture} top) and a RGB Branch (Figure \ref{fig:Network Architecture} bottom). Their output is a pair of feature tensors, \(\textbf{F}_M\) and \(\textbf{F}_I\) respectively, which are fed to the Attention Module to obtain the final feature tensor \(\textbf{F}_A\). This is fed to a linear classifier to obtain the final prediction for a \(K\)-class scene recognition problem.

\subsection{RGB Branch}
It consists of a complete ResNet-18 architecture. This Branch is fed with \(\textbf{I}\) and returns a set of RGB-based features \(\textbf{F}_I \in \mathbb{R}^{w_o \times h_o \times c_o}\) where \(w_o\), \(h_o\) and \(c_o\) are the width, the height and the number of channels of the output feature map. This Branch includes the original ResNet-18's \textit{Basic Blocks} with three of them (\textit{Basic Blocks} 2, 3 and 4) implementing residual connections. \(\textbf{F}_I\), which is obtained from \textit{Basic Block 4}, is then forwarded into the Attention Module.

\subsection{Semantic Branch}
Given \(\textbf{I}\), a semantic segmentation network is used to infer a semantic segmentation score tensor \(\textbf{M}\) which is then fed to the Semantic Branch. \(\textbf{M}\) encodes information of the set of meaningful and representative objects, their spatial relations and their location in the scene. This Branch returns a set of semantic-based features \(\textbf{F}_M \in \mathbb{R}^{w_o \times h_o \times c_o}\) that model the set of semantic labels and their semantic inter-dependencies in spatial and channel dimensions. 

The architecture of the Semantic Branch is a shallow network whose output size exactly matches that of a ResNet-18 feature map (i.e, before the linear classifier). The use of a shallow network is here preferred, as its input \(\textbf{M}\) lacks of texture information. Nonetheless, we compare the effect of using deeper networks for this Branch in section \ref{subsec:Ablation Studies}. 

\begin{figure}[t!]
    \centering
    \includegraphics[width=0.8\linewidth,keepaspectratio]{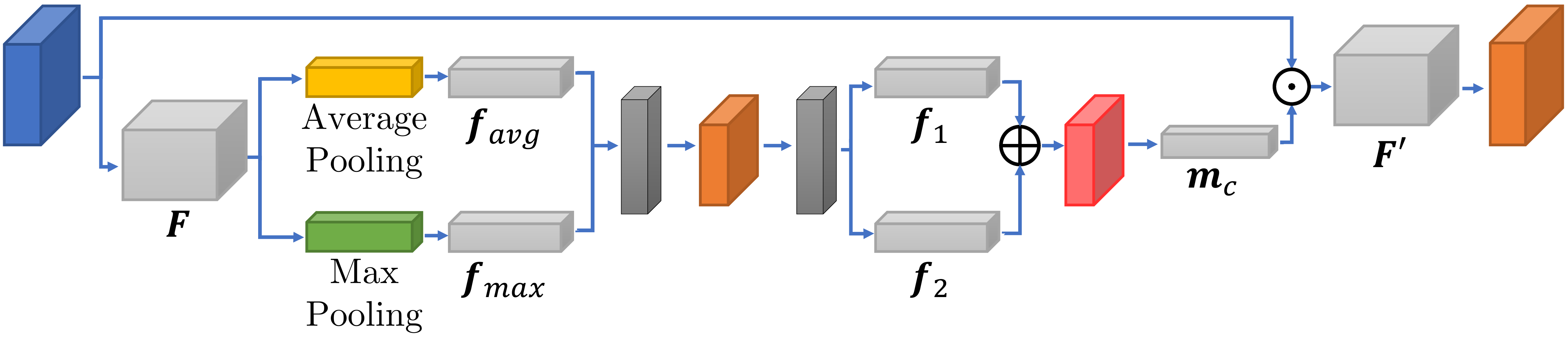}
    \caption{Architecture of a Channel Attention Module (ChAM) \cite{woo2018cbam}. ChAM is fed with an arbitrary size feature map \(\textbf{F} \in \mathbb{R}^{w' \times h' \times c'}\) and infers a 1D channel attention map \(\textbf{m}_c \in \mathbb{R}^{1 \times 1 \times c'}\). Better viewed in color.}
    \label{fig:Channel Attention Module Architecture}
\end{figure}

To reinforce the semantic classes that are relevant for a given image, three Channel Attention Modules (ChAM) \cite{woo2018cbam} are introduced between convolutional blocks in the Semantic Branch. The ChAM module (see Figure \ref{fig:Channel Attention Module Architecture}) exploits the inter-channel relationships between features and results on a per-channel attention map. In this map, some feature channels are reinforced and some inhibited after the sigmoid layer. As each channel represents the probability of a semantic class, ChAM forces the network to focus on certain classes. Specifically, ChAM is fed with an arbitrary size feature map \(\textbf{F} \in \mathbb{R}^{w' \times h' \times c'}\) and infers a 1D channel attention map \(\textbf{m}_c \in \mathbb{R}^{1 \times 1 \times c'}\). \(\textbf{F}\) is squeezed along the spatial dimension by using both average-pooling and max-pooling operations, obtaining feature vectors \(\textbf{f}_{avg}\) and \(\textbf{f}_{max}\). These vectors are combined via two shared fully connected networks separated by a ReLu activation layer. Resulting vectors \(\textbf{f}_{1}\) and \(\textbf{f}_{2}\) are then added and regularized via a sigmoid layer to yield the 1D channel attention map \(\textbf{m}_c\):

\begin{equation}\label{CAM1}
    \begin{split}
        \textbf{f}_1 &= W^{C_2} \cdot \phi ( W^{C_1} \cdot \textbf{f}_{avg} ) \\
        \textbf{f}_2 &= W^{C_2} \cdot \phi ( W^{C_1} \cdot \textbf{f}_{max} ) \\
        \textbf{m}_c(\textbf{F}) &= \sigma (\textbf{f}_1 + \textbf{f}_2),
    \end{split}
\end{equation}

where \(\sigma\) and \(\phi\) are Sigmoid and ReLU activation functions respectively, \(W^{C_1} \in \mathbb{R}^{c/r \times c}\) and \(W^{C_2} \in \mathbb{R}^{c\times r / c}\) are the weights of the fully connected layers, and \(r\) is a reduction ratio parameter.

This channel-attention map is used to weight \(\textbf{F}\) by: 

\begin{equation}\label{CAM2}
        \textbf{F}'= \textbf{m}_c(\textbf{F}) \odot \textbf{F},
\end{equation}
where \((\odot)\) represents a Hadamard product.

\subsection{Attention Module}
\label{subsec:Attention module}
\begin{figure}[t!]
    \centering
    \includegraphics[width=0.5\linewidth,keepaspectratio]{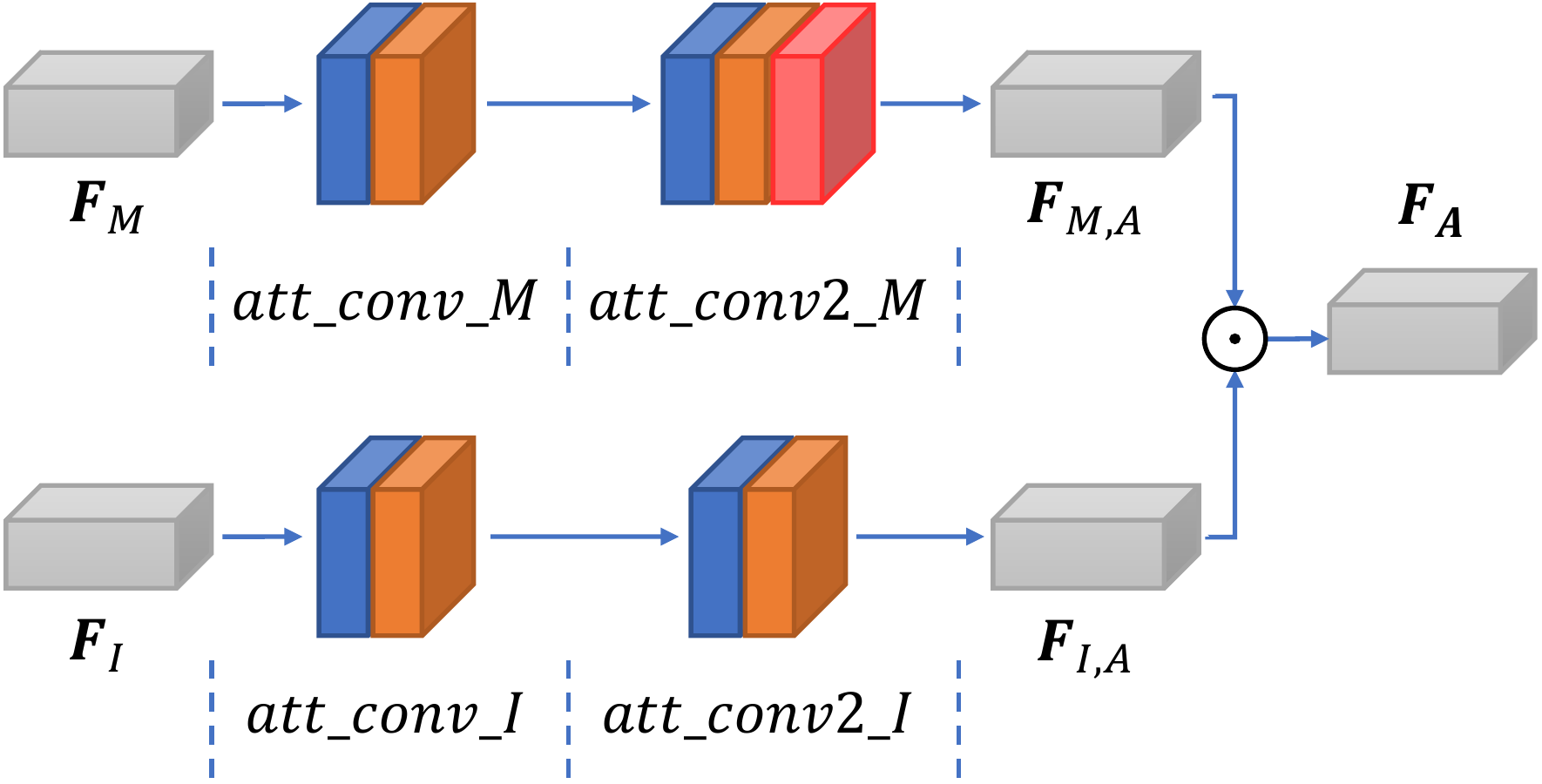}
    \caption{Architecture of the Attention Module. It aims to obtain a set of semantic-weighed features \(\textbf{F}_{A} \in \mathbb{R}^{w_a \times h_a \times c_a}\), based on RGB and semantic representations \(\textbf{F}_{M}\) and \(\textbf{F}_{I}\). Better viewed in color.}
    \label{fig:Attention Module Architecture}
\end{figure}

The Attention Module is used to obtain a set of semantic-weighted features \(\textbf{F}_{A} \in \mathbb{R}^{w_a \times h_a \times c_a}\) which are forwarded to a linear classifier to obtain the final scene posterior probabilities. The architecture and design parameters of the proposed Semantic Attention Module are depicted in Figure \ref{fig:Attention Module Architecture} and detailed in Table \ref{tab:Attention module table summary}. Alternative attention mechanisms have been evaluated (see Section \ref{subsec:Ablation Studies}). Hereinafter, we describe the one that yields the highest performance.

\begin{table}
    \begin{centering}
    \footnotesize
    \renewcommand{\arraystretch}{1.2}
    \centerline{%
        \begin{tabular}{lcc}
            \hline 
            Name & Output Size & Blocks\tabularnewline
            \hline 
            att\_conv\_I & 512 $\times$ 5 $\times$ 5 & 3 $\times$ 3, 512, stride 1\tabularnewline
            att\_conv\_M & 512 $\times$ 5 $\times$ 5 & 3 $\times$ 3, 512, stride 1\tabularnewline
            att\_conv2\_I & 1024 $\times$ 3 $\times$ 3 & 3 $\times$ 3, 1024, stride 1\tabularnewline
            att\_conv2\_M & 1024 $\times$ 3 $\times$ 3 & 3 $\times$ 3, 1024, stride 1\tabularnewline
            attention & 1024 $\times$ 3 $\times$ 3 & Hadamard product \tabularnewline
            avg\_pool & 1024 $\times$ 1 $\times$ 1 & Average Pooling\tabularnewline
            classifier & K $\times$ 1 & Dropout, K-dimensional FC, Softmax\tabularnewline
            \hline 
        \end{tabular}}
        \caption{Layers and output sizes for the Attention Module.}
        \label{tab:Attention module table summary}
    \end{centering}
\end{table}

Semantic gating representations \(\textbf{F}_{M,A}\) are obtained from the output of the Semantic Branch \(\textbf{F}_M\) by a double convolutional block followed by a sigmoid module:

\begin{equation}
    \textbf{F}_{M,A} = \sigma \left( \phi \left( W^{A_2}_M  \left( \phi \left(  W^{A_1}_M \textbf{F}_M + b^{A_1}_M \right)   \right)  + b^{A_2}_M \right)   \right),
\end{equation}

where \(\sigma\) and \(\phi\) are again Sigmoid and ReLU activation functions respectively, \( W^{A_1}_M, b^{A_1}_M \) and \(W^{A_2}_M, b^{A_1}_M\) are the weights and biases of the two convolutional layers. 

Similarly, RGB features to be gated (\(\textbf{F}_{I,A}\)) are obtained from the output of the RGB Branch \(\textbf{F}_I\) by:

\begin{equation}
    \textbf{F}_{I,A} =   \phi \left( W^{A_2}_I  \left( \phi \left(  W^{A_1}_I \textbf{F}_I + b^{A_1}_I \right)   \right)  + b^{A_2}_I \right),
\end{equation}

where \( W^{A_1}_I, b^{A_1}_I \) and \(W^{A_2}_I, b^{A_1}_I\) are the weights of the two convolutional layers of this Branch.

Gating is performed by simply multiplying these two representations:
\begin{equation}
    \textbf{F}_A = 
    \textbf{F}_{I,A} \odot \textbf{F}_{M,A}.
\end{equation}

\(\textbf{F}_A\) is then forwarded to a classifier composed of an average pooling, a dropout, and a fully connected layer, generating a feature vector \(\textbf{f} \in \mathbb{R}^K\).
The inference of scene posterior probabilities \(\textbf{y} \in \mathbb{R}^K\) from \(\textbf{f}\) is achieved by using a logarithmic normalized exponential (\textit{logarithmic softmax}) function \(\gamma(\textbf{f}) : \mathbb{R}^K \rightarrow \mathbb{R}^K\):

\begin{equation}
    \label{eq:softmax}
        y_k = \gamma(\textbf{f}) = \log \left( \frac{\exp(f_k)}{\sum^{}_i \exp(f_i)} \right), \quad \forall i = 1,...,K,
\end{equation}
where \(y_k\) is the posterior probability for class \(k\) given the feature vector \(\textbf{f}\).

\subsection{Training Procedure and Loss} \label{subsec:Training Procedure and Loss}
A two-stage learning procedure has been followed paying particular attention to prevent one of the branches dominating the training process.

Observe that some training examples might be better classified using either RGB or semantic features. In our experiments, when both branches were jointly trained, and one of the branches was more discriminative than the other, the overall loss was small, hindering the optimization of the less discriminative Branch. To avoid this situation, during the first stage, both RGB and Semantic Branches are separately trained for a given scenario (both branches are initialized with Places pretrained weights). At the second training step, trained RGB and semantic branches are frozen while the Attention Module and the linear classifier are fully trained from scratch.

Both training stages are optimized minimizing the following logistic function:
\begin{equation}
    \label{eq:optimization}
       \arg\min_{\psi} \frac{1}{N} \sum^{N}_{i=1} -log(\gamma(\textbf{f})),
\end{equation}
where \(\psi\) denotes the trainable parameters, \(-log(\cdot)\) is the Negative Log-Likelihood (NLL) loss and \(N\) is the number of training samples for a given batch.

\section{Experiments and Results}\label{sec:Experiments and Results}
In this section, we evaluate the proposed scene recognition network on four well-known and publicly available datasets: ADE20K \cite{zhou2017scene}, MIT Indoor 67 \cite{quattoni2009recognizing}, SUN 397 \cite{xiao2010sun} and Places365 \cite{zhou2018places}. To ease the reproducibility of the method, the implementation design of the proposed method is first explained and common evaluation metrics for all the experiments are presented. Then, to assess the effect of each design decision, we perform a varied set of ablation studies. The section closes with a comparison against state-of-the-art approaches on three publicly available datasets.

\begin{figure*}[t!]
    \centering
    \includegraphics[width=1\linewidth,keepaspectratio]{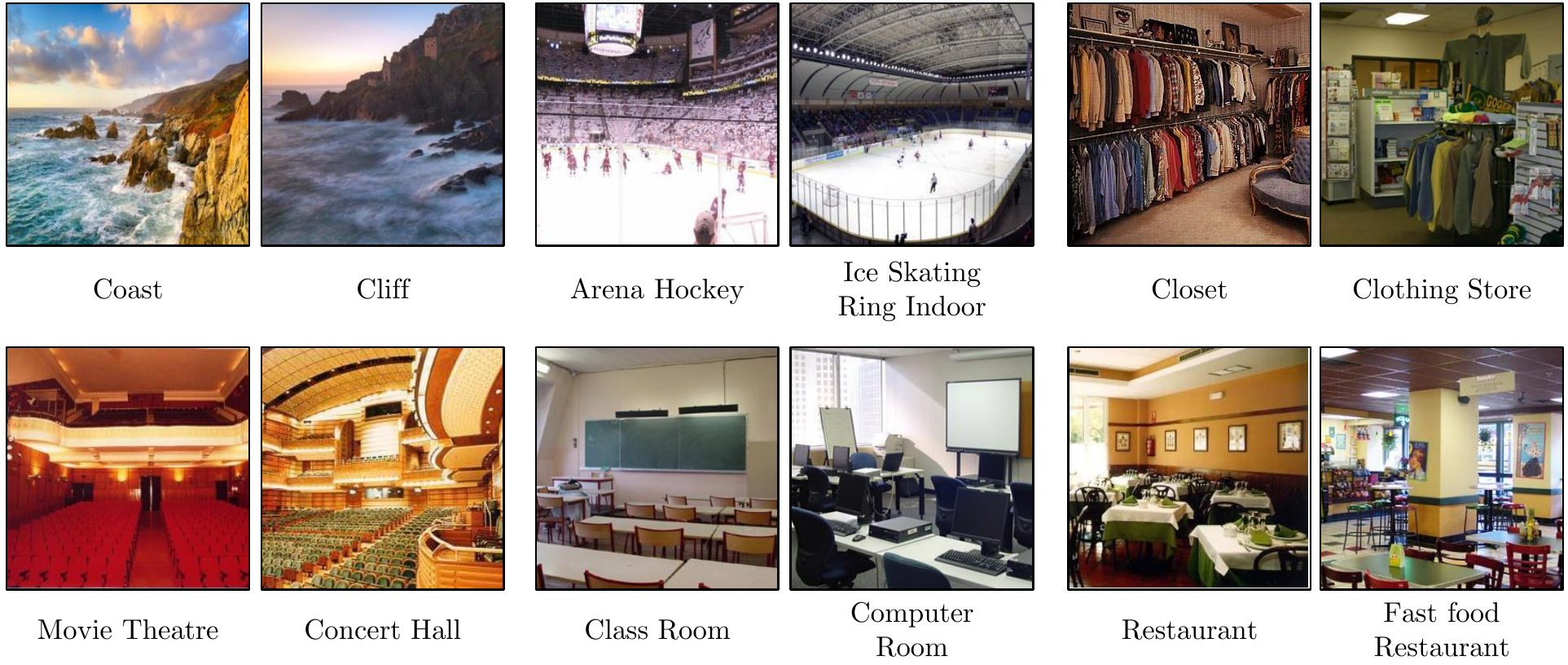}
    \caption{Image examples extracted from ADE20K \cite{zhou2017scene} (top row) and MIT Indoor 67 \cite{quattoni2009recognizing} (bottom row) datasets. Notice the large inter-class similarity between some of the scenes in both datasets. Better viewed in color.}
    \label{fig:DatasetExamples1}
\end{figure*}

\subsection{Implementation Design}\label{Implementation Details}
\textbf{Data Dimensions:} Each input RGB image \(\textbf{I}\) is spatially adapted to the network input by re-sizing the smaller edge to \(256\) and then cropping to a square shape of \(224 \times 224 \times 3\) dimension. In the training stage, this cropping is randomized as part of the data augmentation protocol, whereas in the validation stage, we follow a ten-crop evaluation protocol as described in \cite{zhou2018places}. The Semantic Branch inherits this adaptation. RGB and Semantic Branches produce features \(\textbf{F}_I\) and \(\textbf{F}_M\) respectively, both with dimensions \(512 \times 7 \times 7\). After the Attention Module, a \(1024 \times 3 \times 3\) attention representation \(\textbf{F}_A\) is obtained. This representation is fed to the linear classifier, yielding a probability vector \(y\) with dimensions \(K \times 1\), being \(K\) the number of scene classes of a given dataset.

\textbf{Semantic Segmentation:} It is performed by an UPerNet-50 network \cite{xiao2018unified} fully trained on the ADE20K dataset with \(L=150\) objects and stuff classes. This fixed Semantic Segmentation Network (see Fig. \ref{fig:Network Architecture}) is used throughout these experiments without further adaptation to any dataset.

Its output (a \(150\) vector per pixel, indicating its probability distribution) is the score tensor \(\textbf{M}\). To ease storage and convergence, this tensor is sparsified, i.e. all the \(\textbf{M}_{i,j}\)-th probability distributions which are not among the three top for pixel \((i,j)\) are set to zero.

\textbf{Data Augmentation:} These techniques are used in order to reduce model overfitting and to increase network generalization. Specifically, for the RGB image, we use regular random crop, horizontal flipping, Gaussian blur, contrast normalization, Gaussian noise and bright changes. Due to their nature, for the semantic segmentation tensors, we only apply regular random crop and horizontal flipping. The training dataset is shuffled at each epoch.

\begin{figure*}[t!]
    \centering
    \includegraphics[width=1\linewidth,keepaspectratio]{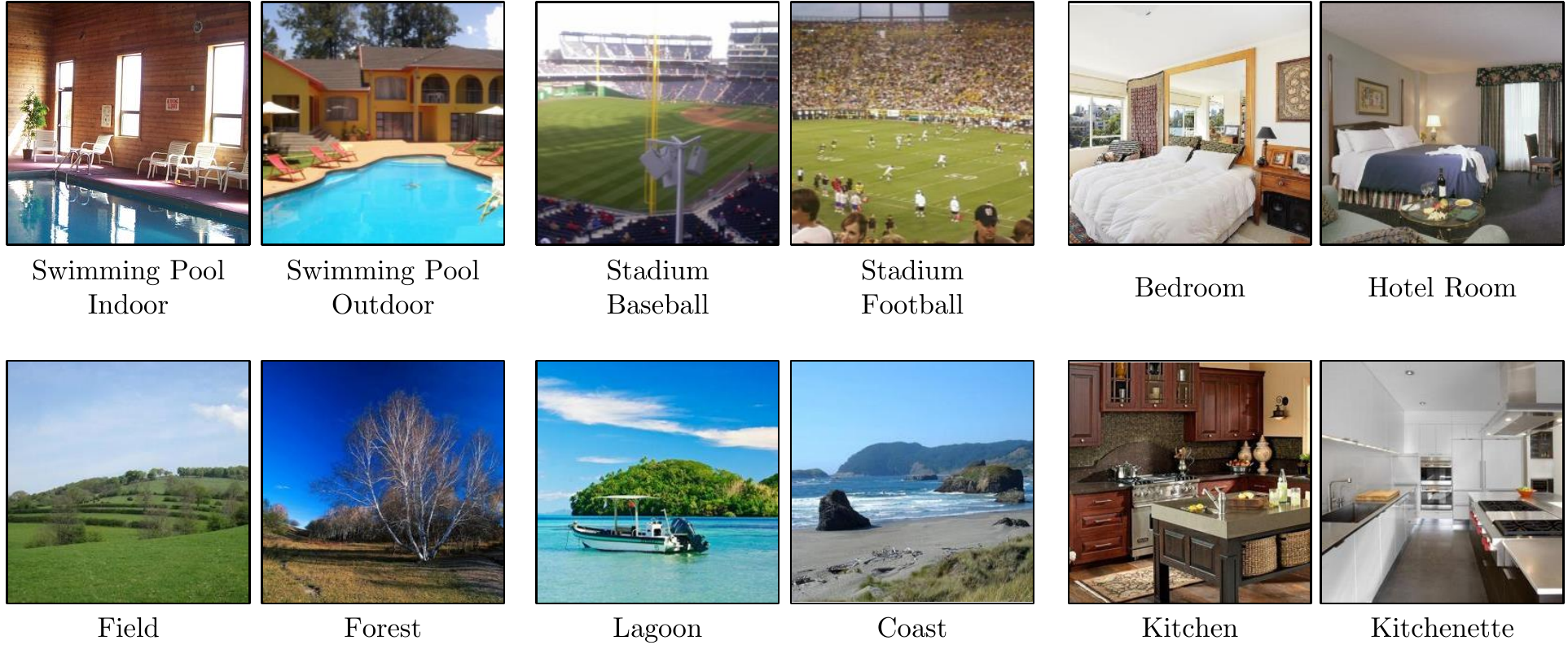}
    \caption{Image examples extracted from SUN 397 \cite{xiao2010sun} (top row) and Places 365 \cite{zhou2018places} (bottom row) datasets. Notice the large inter-class similarity between some of the scenes in both datasets. Better viewed in color.}
    \label{fig:DatasetExamples2}
\end{figure*}

\textbf{Hyper-parameters:} To minimize the loss function in equation \ref{eq:optimization} and optimize the network's trainable parameters \(\psi\), the Deep Frank-Wolfe (DFW) \cite{berrada2018deep} algorithm is used. DFW is a first-order optimization algorithm for deep neural learning which, contrary to the Stochastic Gradient Descend (SGD), only requires the initial learning rate hyper-parameter and not a hand-crafted decay schedule. In all our experiments the initial learning rate was set to \(0.1\). The use of alternative values did not improve learning performance. The use of DFW resulted in similar performance to using SGD with a learning-rate decay schedule suited for each dataset in around the same number of epochs. However, by suppressing the decay policy hyper-parameter, the training process was lightened, as the number of parameters to sweep was simplified. Momentum was set to \(0.9\) and weight decay was set to \(0.0001\) for the two stages in the training procedure (Section \ref{subsec:Training Procedure and Loss}).

\textbf{Hardware and Software:} The model design, training and evaluation has been implemented using PyTorch 1.1.0 Deep Learning framework \cite{paszke2017automatic} running on a PC using a 12 Cores CPU and a NVIDIA TITAN Xp 12GB Graphics Processing Unit. Given this hardware specifications, the proposed architecture increases the training computational time in a \(21.4 \%\) with respect to the RGB baseline: a \(6.78 \%\) due to the training of the semantic branch and a \(14.62 \%\) due to the training of the attention module. When evaluating the proposed network, inference time is increased by a \(37.5 \%\) with respect to the RGB baseline.

\subsection{Datasets} \label{Datasets}
\textbf{ADE20K Scene Parsing Dataset:} The scene parsing dataset ADE20K \cite{zhou2017scene} contains 22210 scene-centric images exhaustively annotated with objects. The dataset is divided into 20210 images for training and 2.000 images for validation. ADE20K is a semantic segmentation benchmark with ground-truth for \(L=150\) semantic categories provided for each image, hence enabling its use for training semantic segmentation algorithms (Section \ref{Implementation Details}-Semantic Segmentation). This dataset is used in our ablation studies (Section \ref{subsec:Ablation Studies}) to reduce the influence of semantic segmentation quality in the evaluation results. Additionally, the dataset includes ground truth for the classification of \(K = 1055\) scene classes. Therefore, it can also be used as a scene recognition benchmark, albeit data composition of scene classes is notably unbalanced. The top row of Figure \ref{fig:DatasetExamples1} depicts example images and scene classes of this dataset.

\textbf{MIT Indoor 67 Dataset:} MIT Indoor 67 dataset \cite{quattoni2009recognizing} contains 15620 RGB images of indoor places arranged onto \(K = 67\) scene classes. In the reported experiments, we followed the division between train and validation sets proposed by the authors \cite{quattoni2009recognizing}: 80 training images and 20 images for validation per scene class. The bottom row of Figure \ref{fig:DatasetExamples1} represents example images and scene classes of the dataset.

\textbf{SUN 397 Dataset:} SUN 397 Dataset \cite{xiao2010sun} is a large-scale scene recognition dataset composed by 39700 RGB images divided into \(K = 397\) scene classes, covering a large variety of environmental scenes both indoor and outdoor. An evaluation protocol to divide the full dataset into both train a validation sets is provided with the dataset \cite{xiao2010sun}. Following such protocol, the dataset is divided into 50 training and validation images per scene class. The top row of Figure \ref{fig:DatasetExamples2} reproduces examples images and scene classes of this dataset.

\textbf{Places 365 Database:} Places365 dataset \cite{zhou2018places} is explicitly designed for scene recognition. It is composed of 10 million images comprising 434 scene classes. There are two versions of the dataset: Places365-Standard with 1.8 million train and 36000 validation images from \(K=365\) scene classes, and Places365-Challenge-2016, in which the size of the training set is increased up to 6.2 million extra images, including 69 new scene classes (leading to a total of 8 million train images from 434 scene classes). In this paper, experiments are carried out using the Places365-Standard dataset. The bottom row of Figure \ref{fig:DatasetExamples2}, presents example images and scene classes of this dataset.

\subsection{Evaluation Metrics}
Scene recognition benchmarks are generally evaluated via the Top@\(k\) accuracy metric with \(k \in [1, K]\). In this paper, Top@\(\lbrace k=1,2,5 \rbrace\) accuracy metrics have been chosen. The Top@\(1\) accuracy measures the percentage of validation/testing images whose top-scored class coincides with the ground-truth label. Top@\(2\) and Top@\(5\) accuracies, are the percentage of validation images whose ground-truth label corresponds to any of the 2 and 5 top-scored classes respectively.

The Top@\(k\) accuracy metrics are biased to classes over-represented in the validation set; or, in other words, under-represented classes barely affect these metrics. In order to cope with unbalanced validation sets (e.g., ADE20K for scene recognition), we propose to use an additional performance metric, the Mean Class Accuracy (MCA):

\begin{equation}
    MCA = \frac{\sum_{i}Top_{i}@1}{K} \quad \forall i=1,...,K,
\end{equation}

where Top\(_{i}\)@\(1\) is the Top@\(1\) metric for scene class \(i\). Note that MCA equals Top@\(1\) for perfectly balanced datasets. 

\begin{table}[!t]
    \begin{centering}
    \footnotesize
    \renewcommand{\arraystretch}{1.2}
    \centerline{%
    \begin{tabular}{lccccccc}
        \hline 
        Backbone & \makecell{Channel Attention \\ Module \cite{woo2018cbam}} & Pretraining & \makecell{Number of \\ Parameters} & Top@1 & Top@2 & Top@5 & MCA\tabularnewline
        \hline 
        ResNet-18 &  & Scratch & 12 M & 49.58 & 60.21 & 71.87 & 11.55\tabularnewline
        ResNet-18 & \checkmark & Scratch & 12 M & 50.60 & 60.45 & 72.10 & 12.17\tabularnewline
        ResNet-18 & \checkmark & ImageNet & 12 M & \textbf{52.17} & \textbf{61.86} & \textbf{71.75} & \textbf{15.54}\tabularnewline
        3 Convolutional Layers &  & Scratch & 6.4 M & 49.80 & 60.55 & 72.53 & 11.37\tabularnewline
        3 Convolutional Layers & \checkmark & Scratch & 6.5 M & 50.00 & 60.95 & 73.53 & 11.67\tabularnewline
        4 Convolutional Layers &  & Scratch & 2.5 M & 49.90 & 60.50 & 71.70 & 12.94\tabularnewline
        4 Convolutional Layers & \checkmark & Scratch & 2.6 M & 50.35 & 60.90 & 72.55 & 13.53 \tabularnewline
        \hline 
    \end{tabular}}
    \caption{Ablation results for different architectures for the Semantic Branch.}
    \label{tab:Ablation Semantic Branch}
    \par
    \end{centering}
\end{table}

\subsection{Ablation Studies}\label{subsec:Ablation Studies}
The aim of this section is to gauge the influence of different design elements that constitute significant aspects of the proposed method. First, the effect of different Semantic Branch architectures is evaluated. Second, the effect of variations in both the attention mechanism and the attention module architecture is assessed and analyzed. Third, the influence of the multi-modal architecture and the attention module with respect to the RGB Branch is quantified. As stated in Section \ref{Datasets}, all the ablation studies are carried out using the ADE20K dataset.

\subsection*{Influence of the Semantic Branch Architecture}
Table \ref{tab:Ablation Semantic Branch} presents results when semantic representations (\(\textbf{F}_M\)) are solely used for scene-recognition. Seven configurations for the Semantic Branch that result from the combination of three design parameters (variation of the network architecture, employment of ChAM modules \cite{woo2018cbam}, and use of pre-trained models) have been explored.

Results from Table \ref{tab:Ablation Semantic Branch} suggest that the inference capabilities of deeper and wider CNNs are not fully exploited, as similar results can be achieved with shallower networks. For instance, when trained from scratch, there is not a significant difference in terms of Top@k and MCA metrics between using ResNet-18 or a shallower network, e.g., a relative improvement of \(0.49\%\) with respect to the 4-convolutional layers configuration. Configurations using 3 or 4 convolutional layers yield similar performance. However, for a given input, a network with 3 convolutional layers (with its associated 3 spatial down-sampling steps) requires larger kernels than a network with 4 convolutional layers (4 down-sampling steps) to obtain a feature output of the same size due to the kernels' size, hence leading to a higher number of parameters. Using a pre-trained network moderately improves performance: ResNet-18 pre-trained on ImageNet performs a \(3.00\%\) (Top@\(1\)) and a \(21.6\%\) (MCA) better than ResNet-18 from scratch, but not drastically, due to the non-RGB nature of the semantic domain. The use of ChAM modules (as in Figure \ref{fig:Network Architecture} for the shallow networks and as in \cite{woo2018cbam} for ResNet-18) slightly improves between a \(2.6\%\) and \(5.4\%\) in terms of MCA, without implying a significant increase in complexity (0.1 Million additional units on average). 

In the light of these experiments, we opted for using the 4 convolutional layer architecture for the Semantic Branch as a trade-off solution between performance and complexity. This architecture yields a comparable performance to ResNet-18 (\(-3.61\%\) and \(-14.86\%\) for Top@\(1\) and MCA respectively), given that substantially less units are used (\(-78.3\%\)).

\begin{figure}[t!]
    \centering
    \includegraphics[width=0.72\linewidth,keepaspectratio]{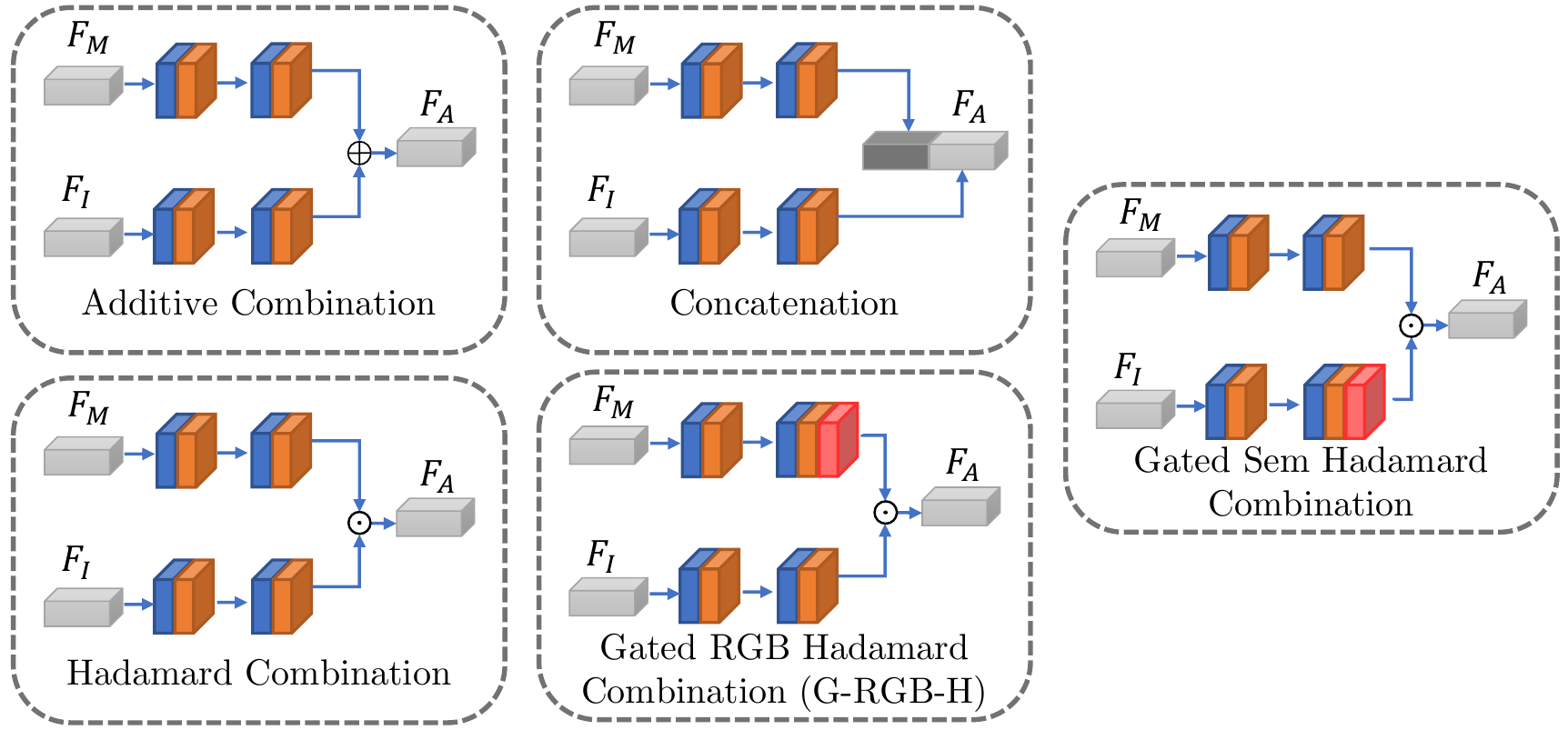}
    \caption{Different attention mechanisms architectures used for the attention module ablation study. Better viewed in color.}
    \label{fig:Attention Mechanisms}
\end{figure}

\subsection*{Influence of the Attention Module}
We have performed two ablation studies on the influence of the attention module. First, we evaluated the effect of using different attention mechanisms all of them sharing a similar network architecture: two convolutional layers for each Branch using \(3\times3\times512\) and \(3\times3\times1024\) kernels. Then, once selected an attention mechanism, we further explore the use of alternative network architectures to drive the attention. 
\subsubsection*{Attention Mechanisms}
Table \ref{tab:Attention Mechanisms} presents an extensive study on attention mechanisms to validate their effectiveness and performance, which is compared with the RGB Baseline. To ease the analysis, Figure \ref{fig:Attention Mechanisms} graphically depicts each of the evaluated attention mechanisms.

Results from Table \ref{tab:Attention Mechanisms} suggest that attention mechanisms generally outperform the RGB baseline in terms of MCA. Additionally, those based on Hadamard combination, either gated or not, perform better than additive and concatenation mechanisms. Additive combination results in a \(4.94\%\) and a \(6.46\%\) decrease in terms of Top@1 and MCA with respect to the Gated RGB Hadamard combination (G-RGB-H). Similarly, attention based on feature concatenation performs a \(1.13\%\) (Top@\(1\)) and a \(6.38\%\) (MCA) worse than G-RGB-H. Besides, concatenation increases the complexity of the linear classifier by generating a larger input to this module. In our opinion, the decreases in performance may be indicative of the inability of additive and concatenation mechanisms to favorable scale scene-relevant information.

Focusing on Hadamard combination, the use of a gated RGB combination (G-RGB-H) slightly improves the non-gated one (a \(4.31\%\) for Top@1 and a \(2.35\%\) for MCA) thanks to the normalized semantic attention obtained after the sigmoid activation layer. The effect of this normalized map is to gate RGB information, hence maintaining its numerical range instead of scaling it as in the non-gated Hadamard combination. Moreover, gating may also serve to nullify not-relevant RGB information, easing the learning process. Additionally, we tested gating attention contrariwise, i.e. using RGB features to gate semantic features (see Figure \ref{fig:Attention Mechanisms} right and Table \ref{tab:Attention Mechanisms} bottom). In this case, feature representations were obtained from the semantic domain, which lacks texture and color information, and the attention map was inferred from RGB domain\textemdash which lacks objects' labels and clear boundaries. These representation problems deteriorate performance a \(9.59\%\) and a \(25.51\%\) for Top@1 and MCA respectively. However, this Gated Sem combination improves the performance of the Semantic Branch itself (compare Table \ref{tab:Ablation Semantic Branch} third row and Table \ref{tab:Attention Mechanisms} last row). In conclusion, we opted for the top-performing G-RGB-H attention mechanism. 

\begin{table}[t!]
    \begin{centering}
    \renewcommand{\arraystretch}{1.2}
    \footnotesize
    \begin{tabular}{lcccc}
        \hline 
        Attention Mechanism & Top@1 & Top@2 & Top@5 & MCA\tabularnewline
        \hline 
        RGB Baseline &  56.90 & 67.25 & 78.00 & 20.80\tabularnewline
        Additive Combination &  59.60 & 70.60 & 80.60 & 25.36\tabularnewline
        Concatenation & 61.85 & 72.25 & 81.60 & 25.38\tabularnewline
        Hadamard Combination & 62.25 & 71.95 & 81.30 & 26.38\tabularnewline
        Gated RGB Hadamard Combination (G-RGB-H) & \textbf{62.55} & \textbf{73.25} & \textbf{82.75} & \textbf{27.00}\tabularnewline
        \hline 
        Gated Sem Hadamard Combination & 56.55 & 66.30 & 75.45 & 20.11\tabularnewline
        \hline 
    \end{tabular}
    \caption{Ablation results for different attention mechanisms.}
    \label{tab:Attention Mechanisms}
    \par
    \end{centering}
\end{table}

\subsubsection*{Architecture of the G-RGB-H Module}
Table \ref{tab:Ablation attention module architectures} presents comparative results for several variations of this module's architecture: straightly gating without adaptation (No Conv layers), using 2 or 3 convolutional layers, and using \(1\times1\) channel-increasing layers.

Results indicate that the inclusion of convolutional layers to \textit{adapt} \(\textbf{F}_I\) and \(\textbf{F}_M\) before the attention mechanism improves performance. The lack of convolutional layers compared with a 2-layer configuration leads to a decrease in performance of a \(2.73\%\) for Top@1 and a \(10.37\%\) for MCA. In our opinion, this owes to the two-stage learning procedure that we follow (see Section \ref{subsec:Training Procedure and Loss}): if no convolutional layers are used, features from the semantic and the RGB Branches are not adapted, hindering the learning process. On the other hand, the use of a deeper attention module (using 3 instead of 2 layers) produces a decrease in performance of \(2.15\%\) and \(7.77\%\) in terms of Top@1 and MCA respectively, suggesting training over-fitting. 

Regarding kernel nature, the use of \(1\times1\) channel-increasing layers improves performance with respect to using no layers at all, but its performance is still behind to that obtained by considering also the width and height dimensions (\(3\times3\) spatial kernels). 

In conclusion, we opted for a G-RGB-H attention strategy based on two convolutional layers composed of \(3\times3\times512\) and \(3\times3\times1024\) kernels at each Branch.

\subsection*{Influence of Semantic Segmentation in Scene Recognition}
Table \ref{tab::Final Results ADE20K} gauges the effectiveness of the proposed architecture when compared to the solely use of either the RGB Branch or the Semantic Branch. Comparative performances show that the RGB and semantic branches highly complement each other when used for scene recognition. The proposed architecture (Table \ref{tab::Final Results ADE20K} last row) outperforms the RGB baseline (Table \ref{tab::Final Results ADE20K} first row): a \(9.92 \%\) in terms of Top@1 performance and a \(29.80 \%\) in terms of MCA. Note that, due to the unbalanced dataset, the MCA metric is more adequate for comparison.

\begin{table}[t!]
    \begin{centering}
    \footnotesize
    \renewcommand{\arraystretch}{1.2}
    \begin{tabular}{cccccc}
        \hline 
        Conv Layers & Kernel Size & Top@1 & Top@2 & Top@5 & MCA\tabularnewline
        \hline 
        \multicolumn{2}{c}{No Conv Layers} & 60.84 & 70.55 & 80.22 & 24.20 \tabularnewline
        \hline 
        \multirow{2}{*}{2} & 1 $\times$ 1 $\times$ 512 & \multirow{2}{*}{\textcolor{black}{61.00}} & \multirow{2}{*}{\textcolor{black}{72.30}} & \multirow{2}{*}{\textcolor{black}{81.95}} & \multirow{2}{*}{\textcolor{black}{24.92}}\tabularnewline
         & 1 $\times$ 1 $\times$ 1024 &  &  &  & \tabularnewline
        \hline 
        \multirow{2}{*}{2} & 3 $\times$ 3 $\times$ 512 & \multirow{2}{*}{\textbf{\textcolor{black}{62.55}}} & \multirow{2}{*}{\textbf{\textcolor{black}{73.25}}} & \multirow{2}{*}{\textbf{\textcolor{black}{82.75}}} & \multirow{2}{*}{\textbf{\textcolor{black}{27.00}}}\tabularnewline
         & 3 $\times$ 3 $\times$ 1024 &  &  &  & \tabularnewline
        \hline 
        \multirow{3}{*}{3} & 3 $\times$ 3 $\times$ 512 & \multirow{3}{*}{\textcolor{black}{61.20}} & \multirow{3}{*}{\textcolor{black}{71.150}} & \multirow{3}{*}{\textcolor{black}{81.750}} & \multirow{3}{*}{\textcolor{black}{24.90}}\tabularnewline
         & 3 $\times$ 3 $\times$ 1024 &  &  &  & \tabularnewline
         & 3 $\times$ 3 $\times$ 1024 &  &  &  & \tabularnewline
        \hline 
    \end{tabular}
    \caption{Ablation results for different G-RGB-H architectures.}
    \label{tab:Ablation attention module architectures}
    \par
    \end{centering}
\end{table} 

\begin{table}[t!]
    \begin{centering}
    \footnotesize
    \renewcommand{\arraystretch}{1.2}
    \begin{tabular}{cccccc}
        \hline 
        RGB & Semantic & Top@1 & Top@2 & Top@5 & MCA\tabularnewline
        \hline 
        \checkmark &  & 56.90 & 67.25 & 78.00 & 20.80\tabularnewline
         & \checkmark & 50.60 & 60.45 & 72.10 & 12.17\tabularnewline
        \checkmark & \checkmark & \textbf{\textcolor{black}{62.55}} & \textbf{\textcolor{black}{73.25}} & \textbf{\textcolor{black}{82.75}} & \textbf{\textcolor{black}{27.00}}\tabularnewline
        \hline 
    \end{tabular}
    \caption{Scene recognition results on ADE20K.}
    \label{tab::Final Results ADE20K}
    \par
    \end{centering}
\end{table}

\begin{figure}[t!]
    \centering
    \includegraphics[width=0.65\linewidth,keepaspectratio]{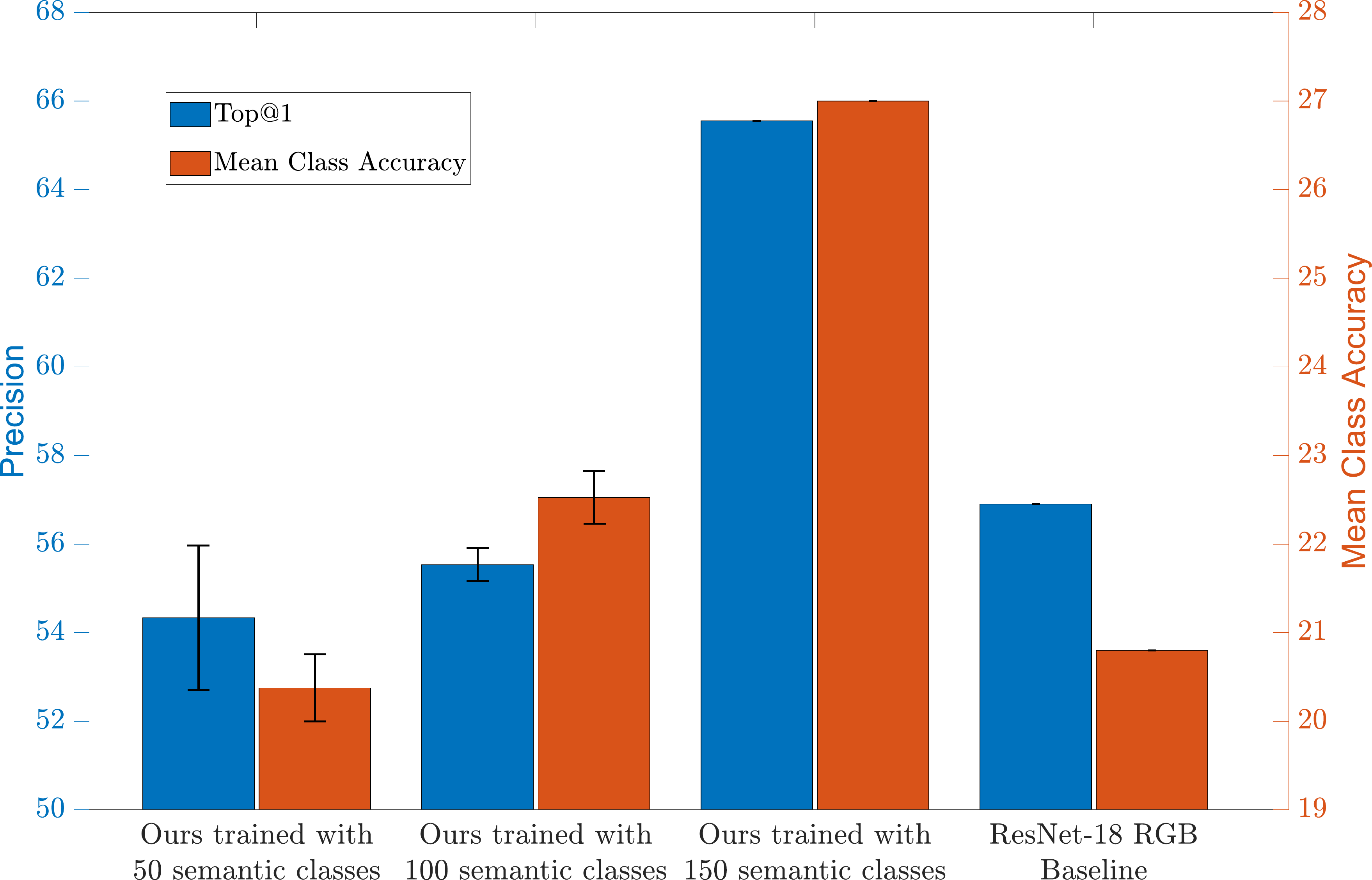}
    \caption{Influence of the number of semantic classes in the performance of the proposed method. For comparison, ResNet-18 RGB baseline is also included in the figure. Average Top@1 metric for the three different random seeds is represented by blue bars and the blue axis. Average Mean Class Accuracy (MCA) is represented by orange bars and the orange axis. Error bars in models with \(L=50\) and \(L=100\) semantic classes are representing standard deviation valuess.}
    \label{fig::Ablation different number of semantic classes}
\end{figure}

In order to extend this result, we have further evaluated the impact of decreasing the number of semantic classes (\(L\), initially 150 classes) in the method's performance. To this aim, two randomly selected sets of semantic classes (\(L=50\) and \(L=100\)) have been used to train the proposed method. The selected sets were incremental, i.e. the set of \(L=50\) semantic classes was contained within the set of \(L=100\) classes. In order to reduce a potential bias of the random selection, three different experiments with three different random seeds have been carried out. For these experiments, Figure \ref{fig::Ablation different number of semantic classes} represents average results by bars' heights and standard deviations via error bars. These indicate that a decrease in the number of semantic classes reduces the performance of the proposed method. If the number of semantic classes is reduced to \(L=100\), Top@1 and MCA metrics reduce by \(15.28 \%\) and by \(16.59 \%\), respectively, compared to considering all classes, but still outperforms the baseline in terms of MCA by a \(8.26 \%\). When the number of semantic classes is further reduced to \(L=50\), performance reduces by \(2.00 \%\) in terms of MCA compared with the ResNet-18 baseline. These results indicate that the performance of the proposed method is reduced if we use few semantic classes, but also suggest that performance might increase if we had more than \(L=150\) semantic classes.

\subsection{State-of-the-art Comparison}
Along this section the proposed method is compared with 19 state-of-the-art approaches, ranging from common CNN architectures trained for scene recognition to methods using objects to drive scene recognition. Comparison is performed on three datasets: MIT Indoor 67 \cite{quattoni2009recognizing}, SUN 397 \cite{xiao2010sun} and Places365 \cite{zhou2018places}. Unless explicitly mentioned, results of all the approaches are extracted from their respective papers.

\begin{table}[t!]
    \begin{centering}
    \renewcommand{\arraystretch}{1.2}
    \footnotesize
    \begin{tabular}{lcccc}
        \hline 
        Method & Backbone & \makecell{Number of \\ Parameters} & Top@1\tabularnewline
        \hline 
        PlaceNet & Places-CNN & $\mathtt{\sim}$ 62 M & 68.24 \tabularnewline
        MOP-CNN & CaffeNet & $\mathtt{\sim}$ 62 M & 68.90\tabularnewline
        CNNaug-SVM & OverFeat & $\mathtt{\sim}$ 145 M & 69.00\tabularnewline
        HybridNet & Places-CNN & $\mathtt{\sim}$ 62 M & 70.80\tabularnewline
        URDL + CNNaug & AlexNet & $\mathtt{\sim}$ 62 M & 71.90\tabularnewline
        MPP-FCR2 (7 scales) & AlexNet & $\mathtt{\sim}$ 62 M & 75.67\tabularnewline
        DSFL + CNN & AlexNet & $\mathtt{\sim}$ 62 M & 76.23\tabularnewline
        MPP + DSF  & AlexNet & $\mathtt{\sim}$ 62 M & 80.78\tabularnewline
        CFV & VGG-19 & $\mathtt{\sim}$ 143 M & 81.00\tabularnewline
        CS & VGG-19 & $\mathtt{\sim}$ 143 M & 82.24\tabularnewline
        SDO (1 scale) \cite{cheng2018scene} & 2\(\times\)VGG-19 & $\mathtt{\sim}$ 276 M & 83.98\tabularnewline
        VSAD \cite{wang2017weakly} & 2\(\times\)VGG-19 & $\mathtt{\sim}$ 276 M & 86.20\tabularnewline
        SDO (9 scales) \cite{cheng2018scene} & 2\(\times\)VGG-19 & $\mathtt{\sim}$ 276 M & 86.76\tabularnewline
        \hline
        RGB Branch & ResNet-18 & $\mathtt{\sim}$ 12 M & 82.68\tabularnewline
        RGB Branch* & ResNet-50 & $\mathtt{\sim}$ 25 M & 84.40\tabularnewline
        Semantic Branch & 4 Conv & $\mathtt{\sim}$ 2.6 M & 73.43\tabularnewline
        Ours & RGB Branch + Sem Branch + G-RGB-H & $\mathtt{\sim}$ 47 M & 85.58\tabularnewline
        \textbf{Ours*} & \textbf{RGB Branch* + Sem Branch} \textbf{+ G-RGB-H} & \textbf{$\mathtt{\sim}$ 85 M} & \textbf{87.10}\tabularnewline
        \hline 
    \end{tabular}
    \caption{State-of-the-art results on MIT Indoor 67 dataset. All stated results with no reference have been extracted from \cite{cheng2018scene}. Methods using objects to drive scene recognition include: \cite{jiang2019deep, cheng2018scene, wang2017weakly}, Semantic Branch, Ours, Ours* and Ours**.}
    \label{tab:MIT Indoor 67 Results}
    \par\end{centering}
\end{table}

\begin{table}[t!]
    \begin{centering}
    \renewcommand{\arraystretch}{1.2}
    \footnotesize
    \begin{tabular}{lccc}
        \hline 
        Method & Backbone & \makecell{Number of \\ Parameters} & Top@1\tabularnewline 
        \hline 
        Decaf & AlexNet & $\mathtt{\sim}$ 62 M & 40.94\tabularnewline
        MOP-CNN  & CaffeNet & $\mathtt{\sim}$ 62 M & 51.98\tabularnewline
        HybridNet & Places-CNN & $\mathtt{\sim}$ 62 M & 53.86\tabularnewline
        Places-CNN & Places-CNN & $\mathtt{\sim}$ 62 M & 54.23\tabularnewline
        Places-CNN ft & Places-CNN & $\mathtt{\sim}$ 62 M & 56.20\tabularnewline
        CS & VGG-19 & $\mathtt{\sim}$ 143 M & 64.53\tabularnewline
        SDO (1 scale) \cite{cheng2018scene} & 2\(\times\)VGG-19 & $\mathtt{\sim}$ 276 M & 66.98\tabularnewline
        VSAD \cite{wang2017weakly} & 2\(\times\)VGG-19 & $\mathtt{\sim}$ 276 M & 73.00\tabularnewline
        SDO (9 scales) \cite{cheng2018scene} & 2\(\times\)VGG-19 & $\mathtt{\sim}$ 276 M & 73.41\tabularnewline
        \hline
        RGB Branch & ResNet-18 & $\mathtt{\sim}$ 12 M & 67.65\tabularnewline
        RGB Branch* & ResNet-50 & $\mathtt{\sim}$ 25 M & 70.87\tabularnewline
        Semantic Branch & 4 Conv & $\mathtt{\sim}$ 2.6 M & 51.32\tabularnewline
        Ours &  RGB Branch + Sem Branch + G-RGB-H & $\mathtt{\sim}$ 47 M & 71.25\tabularnewline
        \textbf{Ours*} & \textbf{RGB Branch* + Sem Branch} \textbf{+ G-RGB-H} & \textbf{$\mathtt{\sim}$ 85 M} & \textbf{74.04}\tabularnewline
        \hline 
    \end{tabular}
    \caption{State-of-the-art results on SUN 397 dataset. All stated results with no reference have been extracted from \cite{cheng2018scene}. Methods using objects to drive scene recognition include: \cite{cheng2018scene, wang2017weakly}, Semantic Branch, Ours, Ours* and Ours**.}
    \label{tab:SUN 397 Results}
    \par\end{centering}
\end{table}

\subsection*{Evaluation on MIT Indoor 67 and SUN 397}
Tables \ref{tab:MIT Indoor 67 Results} and \ref{tab:SUN 397 Results} agglutinate performances of scene recognition methods for the MIT Indoor 67 and the SUN 397 datasets respectively. As these datasets are balanced, we only include the Top@1 metric. All the compared algorithms are based on CNNs (see the Backbone column for details). For this evaluation, and just to offer a fair comparison in terms of network complexity, we also present results for a version of the proposed method (Ours) using a ResNet-50 network, instead of a ResNet-18 one, for the RGB Branch (Ours*). Note that, ResNet-50 is preferred over a higher-capacity network as VGG-19. The reason for this choice lies on the distribution of the discriminative power in each architecture. Regarding only the learnable layers, VGG-19 is composed of 16 convolutional layers, followed by a powerful classification module made up of three fully connected layers that constitute around \(86 \%\) of the learnable capacity of the network. Consequently, during the training process a relevant fraction of the discriminative power is learned by the classifier. In our experiments, we noticed that this unbalanced distribution conveys features of a lower interpretability at the output of the convolutional layers that, in turn, are less useful for the proposed scheme. Differently, the convolutional layers of ResNet-50 represent around \(97 \%\) of the learnable capacity of the network, providing a high interpretability power to the features entering the attention module. The results of these experiments agree with the neuron interpretability profiles discussed on \cite{bau2017network}.

Results on MIT Indoor 67 dataset (see Table \ref{tab:MIT Indoor 67 Results}) are in line with those of Table \ref{tab::Final Results ADE20K}, suggesting a high complementarity between the RGB and the semantic representations. The proposed method increases Top@1 performance of the RGB Branch by a \(3.50 \%\) and a \(3.19 \%\) for ResNet-18 and ResNet-50 backbones respectively. These relative increments are smaller than those reported in Table \ref{tab::Final Results ADE20K} as the semantic segmentation model, trained with the ADE20K dataset, is less tailored to the MIT scenarios. The proposed method (Ours) performs better than most of state-of-the-art algorithms while using a substantially smaller number of parameters. e.g., a \(1.90 \%\) improvement over the single scale SDO \cite{cheng2018scene} (a method similar in spirit) while reducing a \(82.97 \%\) the number of parameters. The ResNet-50 backbone version of our method (Ours*) outperforms every other state-of-the-art method, providing relative performance increments of \(1.10 \%\) and \(0.40 \%\) with respect to VSAD and multi-scale SDO, still with a significant reduction in the model's complexity\textemdash \(69.20\%\) less parameters. Additionally, whereas multi-scale patch-based algorithms require severe parametrization of the size, stride and scale of each of the explored patches, the proposed algorithm is only parametrized in terms of training hyper-parameters, highly simplifying its learning process.

A similar discussion applies to the SUN 397 Dataset (see Table \ref{tab:SUN 397 Results}). The proposed method (Ours) outperforms most of state-of-the-art approaches, maintaining lower complexity; and if we go for a more complex network for the RGB Branch (Ours*), the proposed method outperforms all reported methods still using less than 1/3 of parameters.

\begin{table}[t!]
    \begin{centering}
    \footnotesize
    \renewcommand{\arraystretch}{1.2}
    \begin{tabular}{lccccc}
        \hline 
        Network & \makecell{Number of \\ Parameters} & Top@1 & Top@2 & Top@5 & MCA\tabularnewline
        \hline 
        AlexNet & $\mathtt{\sim}$ 62 M & 47.45 & 62.33 & 78.39 & 49.15\tabularnewline
        AlexNet {{@\cite{zhou2018places}}} & $\mathtt{\sim}$ 62 M & 53.17 & - & 82.89 & -\tabularnewline
        GoogLeNet {{@\cite{zhou2018places}}} & $\mathtt{\sim}$ 7 M & 53.63 & - & 83.88 & -\tabularnewline
        ResNet-18 & $\mathtt{\sim}$ 12 M & 53.05 & 68.87 & 83.86 & 54.40\tabularnewline
        ResNet-50 & $\mathtt{\sim}$ 25 M & 55.47 & 70.40 & 85.36 & 55.47\tabularnewline
        ResNet-50 {{@\cite{zhou2018places}}} & $\mathtt{\sim}$ 25 M & 54.74 & - & 85.08 & -\tabularnewline
        VGG-19 {{@\cite{zhou2018places}}} & $\mathtt{\sim}$ 143 M & 55.24 & - & 84.91 & -\tabularnewline
        DenseNet-161 & $\mathtt{\sim}$ 29 M & 56.12 & 71.48 & 86.12 & 56.12\tabularnewline
        \hline 
        Semantic Branch & $\mathtt{\sim}$ 2.6 M & 36.20 & 50.11 & 68.48 & 36.20\tabularnewline
        \textbf{Ours} & \textbf{$\mathtt{\sim}$ 47 M} & \textbf{56.51} & \textbf{71.57} & \textbf{86.00} & \textbf{56.51}\tabularnewline
        \hline 
    \end{tabular}
    \caption{State-of-the-art results on Places-365 Dataset (\%). ({@\cite{zhou2018places}} stands for performance metrics reported in \cite{zhou2018places}).}
    \label{tab:Places365 Results}
    \par\end{centering}
\end{table}

\begin{figure*}[t!]
    \centering
    \includegraphics[width=1\linewidth,keepaspectratio]{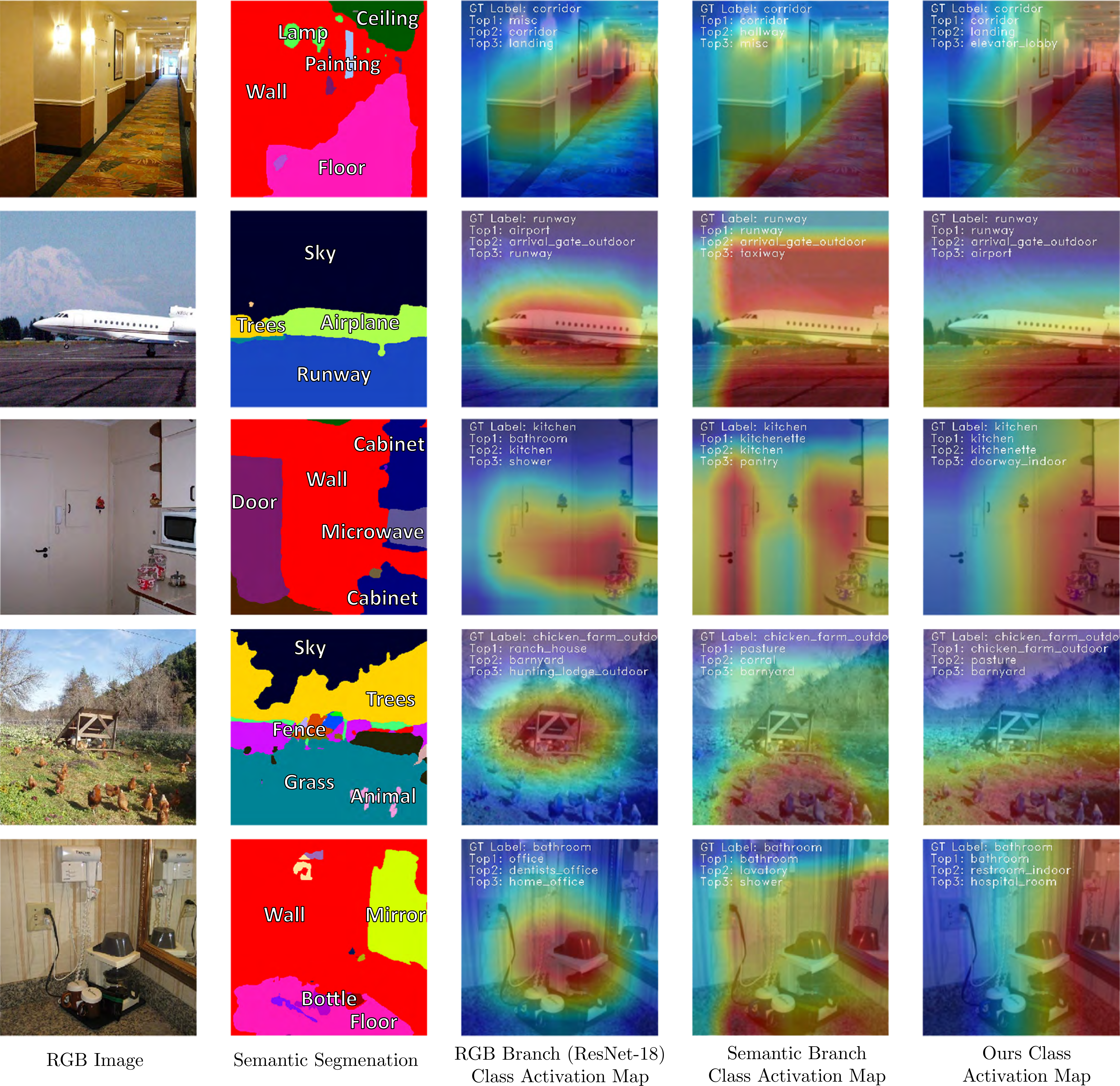}
    \caption{Qualitative results. First and second column represent the RGB and semantic segmentation images for selected examples of the ADE20K, the SUN 397 and the Places 365 validation sets. The third, fourth and fifth columns depict the Class Activation Map (CAM) \cite{zhou2015cnnlocalization} obtained by using features extracted from: the RGB Branch used baseline (ResNet-18), the Semantic Branch and the proposed method (Ours). CAM represents the image areas that produce a greater activation of the network (the redder the larger). CAM images also indicate the ground-truth label and the Top 3 predictions. Better viewed in color.}
    \label{fig:Qualitvate Attention Maps}
\end{figure*}

\subsection*{Evaluation on Places 365}
Table \ref{tab:Places365 Results} presents results for the challenging Places365 Dataset. This Table includes both results from the original Places365 paper (@\cite{zhou2018places} in Table \ref{tab:Places365 Results}) and those we obtained after the downloading and evaluation of the publicly available network models. 

The proposed method (Ours) obtains the best results while maintaining relatively low complexity. Its performance improves those of the deepest network, DenseNet-161, by a \(0.73 \%\) in terms of Top@1 accuracy and it surpasses the highest-capacity network, VGG-19, by a \(2.29 \%\) reducing the number of parameters a \(67.13 \%\).

\subsection{Qualitative Interpretation of the Attention Module}
\textbf{Class Activation Maps:}
To qualitatively analyze the benefits of the attention module, Figure \ref{fig:Qualitvate Attention Maps} depicts intermediate results of our method, including, from left to right: the original color image, its semantic segmentation and Class Activation Maps (CAMs) \cite{zhou2015cnnlocalization} at the output of the RGB Branch, through the \(\textbf{F}_I\) features, the Semantic Branch, via the \(\textbf{F}_M\) features, and after the attention module, using the \(\textbf{F}_A\) features. CAMs are here represented normalized in color with dark-red and dark-blue representing maximum (1) and minimum (0) activation respectively. Top 3 predicted classes for each Branch are included at the top-left corner of each image.

The automatic refocusing capability of the attention module can be observed at first glance. Whereas CAMs of the RGB Branch are clearly biased towards the image center, after the attention module, the attention is focused on human-accountable concepts that can be indicative of the scene class, e.g., the microwave for the kitchen, the animals for the chicken farm or the mirror for the bathroom. This refocusing is specially useful for the disambiguation between similar classes\textemdash e.g., the runway to correct the airport prediction at the second row, and to drive attention towards discriminative objects in conflicting scenes\textemdash e.g., the mirror can be used to recognize the bathroom at the last row.

Owing to the trained convolutional layers in the attention module (see Section \ref{subsec:Attention module}) the final CAM is not derived from the straight multiplication of the RGB and the Semantic CAMs. Instead, the effect of the Semantic CAM\textemdash together with refocusing, is generally an enlargement of the focus area (compare third and fifth columns in the Figure \ref{fig:Qualitvate Attention Maps}); hence, increasing the amount of information that may be used to discriminate between scene classes. This may be a consequence of the larger CAMs yielded by the Semantic Branch: as the semantic segmentation contains less information than the color image, the Semantic Branch tends to focus on either small discriminative objects or on large areas containing objects' transitions.

\begin{figure*}[t!]
    \centering
    \includegraphics[width=1\linewidth,keepaspectratio]{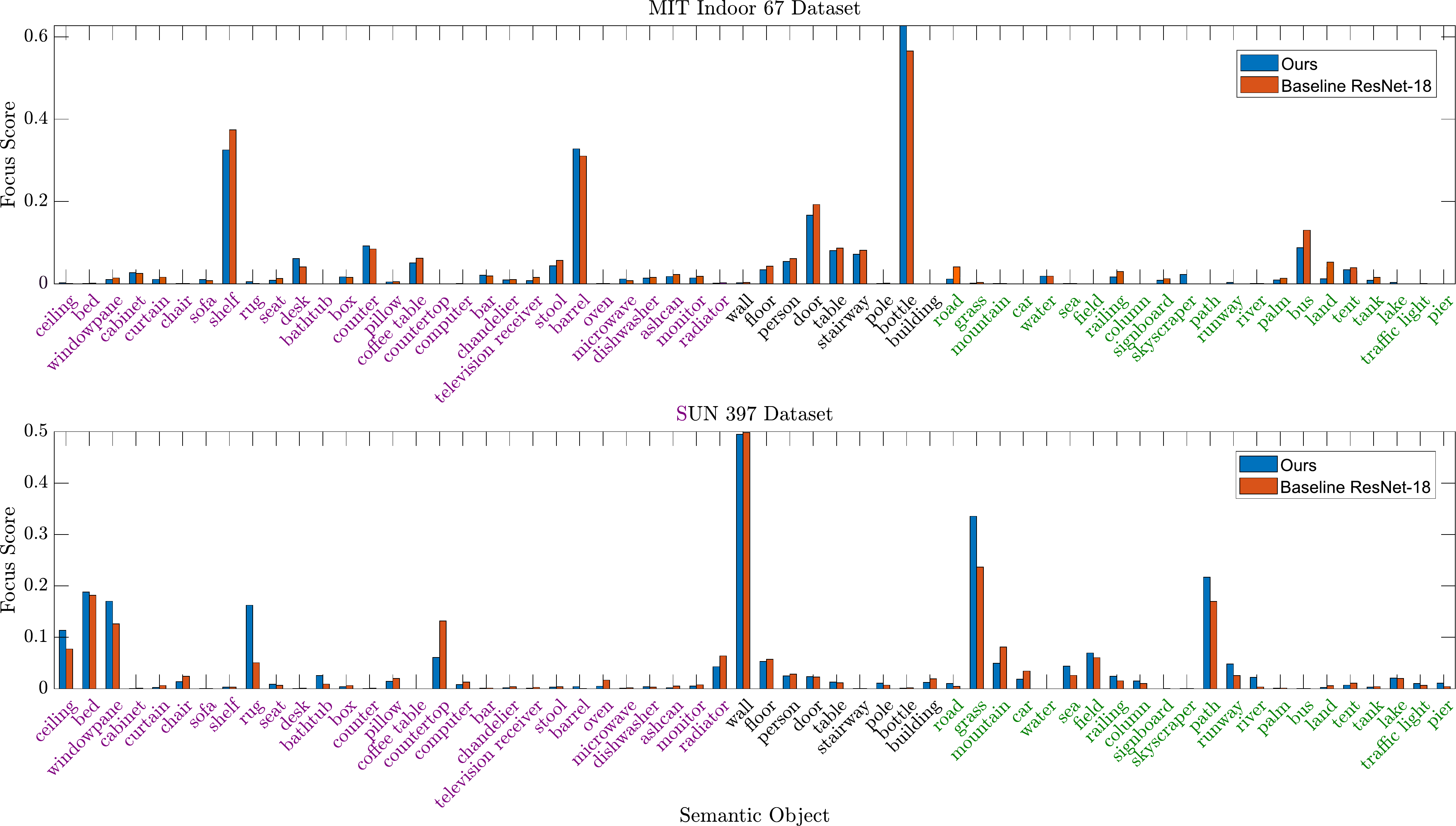}
    \caption{Correlation between objects and scene concepts in both MIT Indoor 67 (top bar graph) and SUN 397 datasets (bottom bar graph) for ResNet-18 Baseline and for the proposed method (Ours). Objects/stuff are divided into 3 separate subsets. Purple color is used to represent objects that are \textit{Indoors}, green objects are \textit{Outdoors}, and black ones are those that lie in \textit{Both} indoors and outdoors. Better viewed in color.}
    \label{fig:Network objects focus}
\end{figure*}

\textbf{Correlation between scene and objects:}
Using the Class Activation Maps described in the previous section we can obtain the correlation between objects and scene concepts, by  taking into account the activation received by each object/stuff class in both the MIT Indoor 67 and the SUN 397 datasets. Accumulated attentions along the dataset are depicted in Figure \ref{fig:Network objects focus}. The horizontal axis represents the set of semantic classes assigned a larger focus---according to the CAMs---for both datasets. Objects/stuff are divided into 3 separate subsets: purple color is used to represent objects that are common in \textit{Indoor} scenarios, green objects are those that can be found in \textit{Outdoor} scenarios, and black ones are those that may be found in \textit{Both} scenarios.

The top graph in Figure \ref{fig:Network objects focus} suggests that, for MIT Indoor 67 dataset, the proposed network (blue bars) is essentially focused on stuff/objects associated with the \textit{Indoor} (purple) and \textit{Both} (black) subsets. Comparing with the Baseline (orange bars), the attention received by \textit{Outdoor} semantic classes decreases, e.g. see  ``road'', ``railing'', ``bus'' or ``land''. These results are consistent with the MIT Indoor dataset which only includes indoor scenarios. Moreover, the correlation between objects and scene concepts  for the SUN 397 dataset (Figure \ref{fig:Network objects focus} bottom graph) suggests a more evenly distributed attention towards stuff/objects from \textit{Indoor}, \textit{Outdoor} and \textit{Both} sets---disregarding objects' sizes.

Results for MIT Indoor and SUN 397 datasets suggest that the most attended semantic classes depend on whether the dataset's scenes are from \textit{Indoor}, \textit{Outdoor} or \textit{Both} scenarios. We believe that this preliminary analysis sets a base for forthcoming interpretability studies that will be part of our future work.

\begin{figure*}[t!]
    \centering
    \includegraphics[width=1\linewidth,keepaspectratio]{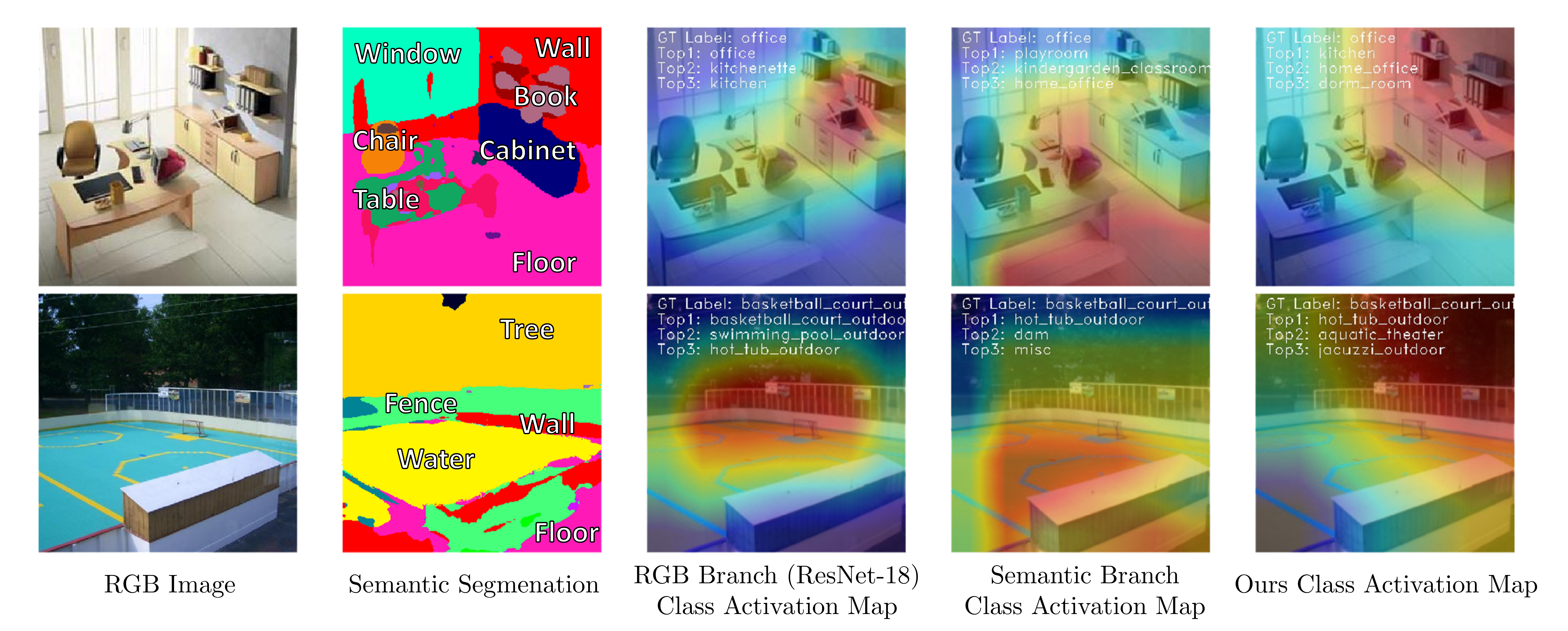}
    \caption{Examples of limitations of the proposed method. In some cases, when semantic segmentation is insufficient (top row) or incorrect (bottom row) network predictions are erroneous. In the top row, the presence of a cabinet and the absence of the computer in the semantic segmentation leads to a kitchen prediction. In the bottom row, the erroneous segmentation of the floor as water leads to water-related scene predictions.}
    \label{fig:Model Limitations}
\end{figure*}

\subsection{Limitations of the Proposed Method}
According to the reported results, semantic segmentation is useful to gate the process of scene recognition on RGB images. However, if semantic segmentation is flawed or imprecise, the proposed method may not be able to surpass the erroneous information.

Figure \ref{fig:Model Limitations} depicts two qualitative examples of these problems. In the top row, the semantic image lacks information on discriminative objects e.g., the \textit{computer}. In the absence of these objects, the \textit{cabinet} object, which is correctly segmented, dominates the gating process and drives an erroneous recognition: a "Kitchen" instead of an "Office". The bottom row contains another problematic situation. The semantic segmentator miss-classifies the court as \textit{water} (probably due to its color and texture). The proposed network, guided by the primary presence of \textit{water}, infers that the 3 most probable scenes classes are those in which water is preeminent, hence failing.

\section{Conclusions}\label{sec:Conclusions}
This paper describes a novel approach to scene recognition based on an end-to-end multi-modal convolutional neural network. The proposed method gathers both image and context information using a two-branched CNN (the traditional RGB branch and a complementary semantic information branch) whose features are combined via an attention module. Several attention strategies have been explored including classical additive and concatenating strategies as well as novel strategies based on fully trained convolutional layers followed by a Hadamard product. Among these last, the top performing strategy relies on a softmax transformation of the convolved Semantic Branch features. These transformed features are used to gate convolved features of the RGB Branch, which results in the reinforcement of the learning of relevant context information by changing the focus of attention towards human-accountable concepts indicative of scene classes. 

Results on publicly available datasets (ADE20K, MIT Indoor 67, SUN 397 and Places 365) confirm that the proposed method outperforms every reported state-of-the-art method while significantly reducing the number of network parameters. This model simplification decreases training and inference times and the amount of necessary data for the learning stage. Overall, results confirm that the combination of both RGB and semantic segmentation modalities generally benefits scene recognition.

\section*{Acknowledgment}
This study has been partially supported by the Spanish Government through its TEC2017-88169-R MobiNetVideo project.

\section*{References}

\bibliography{mybibfile}

\end{document}